\documentclass[twoside]{article}
\usepackage{amsfonts,amssymb,amsbsy,textcomp,marvosym,amsmath,caption,threeparttable,amsthm,subfigure,float,lastpage,lscape}
\usepackage{eurosym,mathrsfs,fancyhdr,CJK,multicol,graphics,indentfirst,color,bm,upgreek,booktabs,graphicx,multirow,warpcol}
\usepackage{epstopdf}

\usepackage[ruled,vlined]{algorithm2e}

\usepackage{xr-hyper}
\usepackage{varioref}
\PassOptionsToPackage{hyphens}{url}
\usepackage{cite}
\makeatletter
\let\NAT@parse\undefined
\makeatother
\usepackage[colorlinks,linkcolor = blue,anchorcolor = blue,citecolor = blue,urlcolor = black]{hyperref}
 
\usepackage[capitalise,nameinlink,]{cleveref}

\newtheorem{theorem}{\hskip 1em Theorem}

\theoremstyle{remark}
\newtheorem{remark}{\hskip 1em Remark}

\theoremstyle{definition}

\newtheorem{assumption}{\hskip 1em Assumption}
\def\footnoterule{\kern 1mm \hrule width 10cm \kern 2mm}

\allowdisplaybreaks
\catcode`@=11
\def\title#1{\vspace{3mm}\begin{flushleft}\vglue-.1cm\Large\bf\boldmath\protect\baselineskip=18pt plus.2pt minus.1pt #1
\end{flushleft}\vspace{1mm} }
\def\author#1{\begin{flushleft}\normalsize #1\end{flushleft}\vspace*{-4pt} \vspace{3mm}}
\def\address#1#2{\begin{flushleft}\vglue-.35cm${}^{#1}$\small\it #2\vglue-.35cm\end{flushleft}\vspace{-2mm}\par}
\def\jz#1#2{{$^{\footnotesize\textcircled{\tiny #1}}$\let\thefootnote\relax\footnotetext{\!\!$^{\footnotesize\textcircled{\tiny #1}}$#2}}}
\catcode`@=11
\def\section{\@startsection{section}{1}{\z@}%
 {-3ex \@plus -.3ex \@minus -.2ex}%
 {2.2ex \@plus.2ex}%
{\normalfont\normalsize\protect\baselineskip=14.5pt plus.2pt minus.2pt\bfseries}}
\def\subsection{\@startsection{subsection}{2}{\z@}%
 {-3ex\@plus -.2ex \@minus -.2ex}%
 {2ex \@plus.2ex}%
{\normalfont\normalsize\protect\baselineskip=12.5pt plus.2pt minus.2pt\bfseries}}
\def\subsubsection{\@startsection{subsubsection}{3}{\z@}%
 {-2.2ex\@plus -.21ex \@minus -.2ex}%
 {1.4ex \@plus.2ex}
{\normalfont\normalsize\protect\baselineskip=12pt plus.2pt minus.2pt\sl}}

\begin{document}
\begin{CJK*}{GBK}{song}
\thispagestyle{empty}
\vspace*{-13mm}
\vspace*{2mm}

\title{FedMeS: Personalized Federated Continual Learning Leveraging Local Memory}

\author{Jin Xie$^{1}$, Chenqing Zhu$^{1}$, and Songze Li$^{2}$}

\address{1}{The Hong Kong University of Science and Technology (Guangzhou)}
\address{2}{Southeast University}

\let\thefootnote\relax\footnotetext{{}\\[-4mm]\indent\ Regular Paper}

\noindent {\small\bf Abstract} \quad  We focus on the problem of Personalized Federated Continual Learning (PFCL): a group of distributed clients, each with a sequence of local tasks on arbitrary data distributions, collaborate through a central server to train a personalized model at each client, with the model expected to achieve good performance on all local tasks. We propose a novel PFCL framework called Federated Memory Strengthening ({\ttfamily FedMeS}) to address the challenges of client drift and catastrophic forgetting. In {\ttfamily FedMeS}, each client stores samples from previous tasks using a small amount of local memory, and leverages this information to both 1) calibrate gradient updates in training process; and 2) perform KNN-based Gaussian inference to facilitate personalization. {\ttfamily FedMeS} is designed to be \emph{task-oblivious}, such that the same inference process is applied to samples from all tasks to achieve good performance. {\ttfamily FedMeS} is analyzed theoretically and evaluated experimentally. It is shown to outperform all baselines in average accuracy and forgetting rate, over various combinations of datasets, task distributions, and client numbers.

\vspace*{3mm}


\vspace*{4mm}

\end{CJK*}
\baselineskip=18pt plus.2pt minus.2pt
\parskip=0pt plus.2pt minus0.2pt
\begin{multicols}{2}

\section{Introduction}

Federated learning (FL) \cite{mcmahan2017communication} is an emerging distributed learning framework that allows for collaborative training of a model across multiple clients while keeping their raw data locally stored. 
A typical FL process involves local training on each client and global model aggregation on a cloud server, with only model updates or gradients being shared 
between clients and the server.

\begin{figure}[H]
\begin{center}
\centerline{\includegraphics[width=8cm]{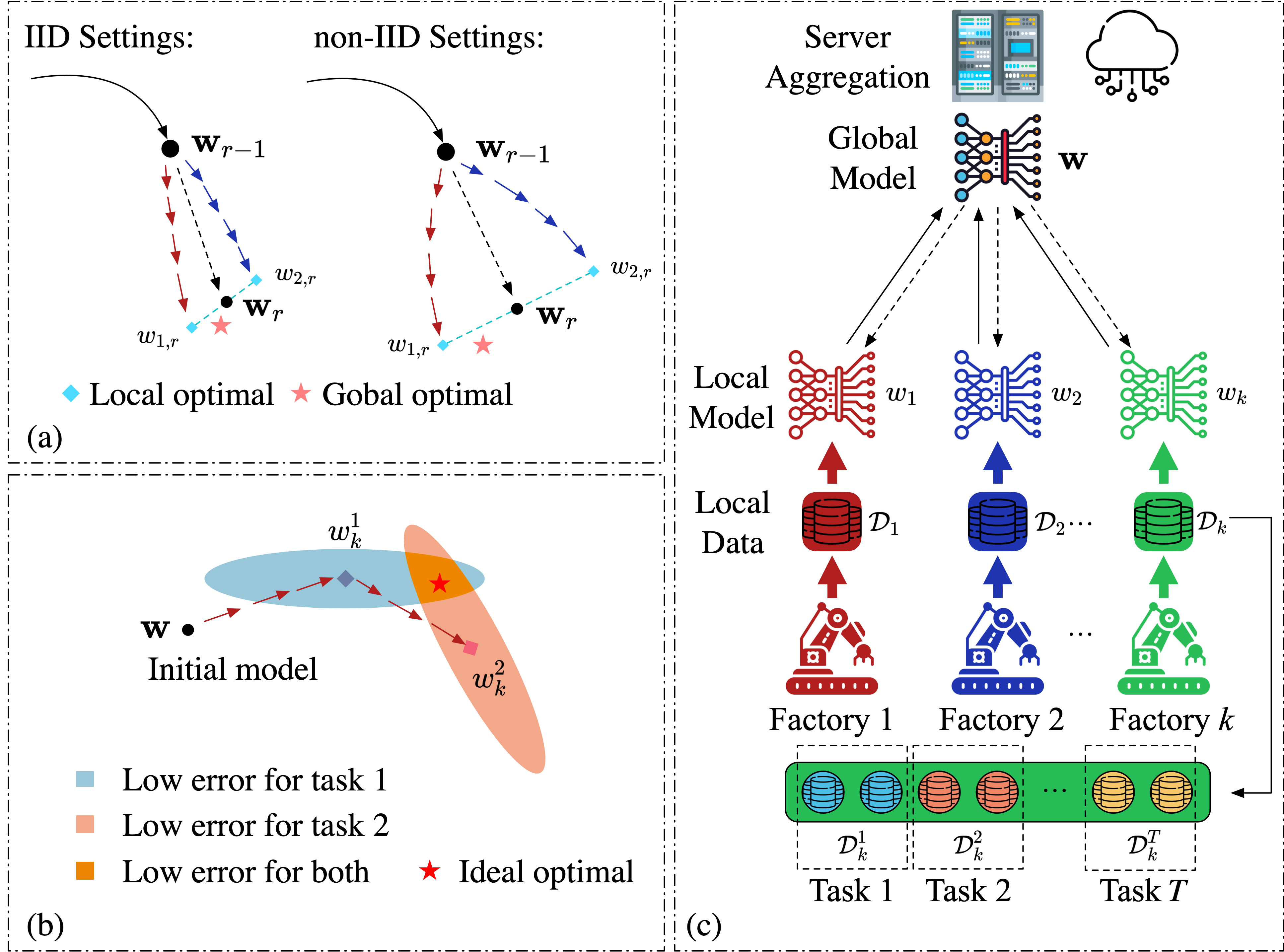}}
\caption{(a) Illustration of the model parameter divergence with non-IID datasets. (b) Illustration of catastrophic forgetting. (c) An overview of a PFCL system in IIoT scenario.}
\label{fig1}
\end{center}
\vspace{-8mm}
\end{figure}

Data collected from different clients in an FL system often have drastically different distributions. As seen in Figure \cref{fig1}(a), this can lead to model parameter divergence and \textit{client drift} \cite{venkatesha2022addressing}, causing potentially poor performance for certain clients. The conventional way of training a single model is insufficient to fit all the non-IID data, and a personalized model needs to be trained for each participating client~\cite{huang2021personalized,fallah2020personalized}, which is known as personalized FL. 

Another key characteristic in real-world FL systems is that clients are continuously collecting new data (new task) which may exhibit different distributions from previous local data (tasks). Hence, it would be preferable to train a local model that is able to achieve consistently good performance in all local tasks. In an FL system, the problem is solved by federated continual learning (FCL) \cite{yao2020continual,luopan2022fedknow,yoon2021federated}. The phenomenon of a model failing to perform well on previously trained tasks is called \textit{catastrophic forgetting} \cite{kirkpatrick2017overcoming}, which is illustrated in \cref{fig1}(b).

In practical FL systems, the data and task heterogeneity often exist both across clients and over time on a single client. For instance, as shown in \cref{fig1}(c), in a IIoT scenario, multiple factories manufacturing different products would like to use FL for training defect detection models collaboratively. Other than the difference between the types of products, each factory may experience change of tasks over time due to e.g., change of recipe and upgrade of the production line. Aiming to address the challenges of client drift and catastrophic forgetting \emph{simultaneously}, in the paper we focus on the personalized federated continual learning (PFCL) problem. In a PFCL system, each client observes a stream of arbitrarily different tasks, and would like to collaborate through the server to train a personalized model, which performs well on both previous and current local tasks.


There have been prior attempts at tackling the FCL problem, like {\ttfamily FedWeIT} \cite{yoon2021federated}, where task-generic and task-specific knowledge are shared across clients to decompose the model parameters. However, as demonstrated in \cite{venkatesha2022addressing}, {\ttfamily FedWeIT} struggles to handle the issue of client drift caused by data heterogeneity and has poor scalability as the number of tasks increases.

In this paper, we propose a novel PFCL framework called \textbf{F}ederated \textbf{M}emory \textbf{S}trengthening ({\ttfamily FedMeS}). 
{\ttfamily FedMeS} utilizes small amount of local memory at each client to store information about previous tasks, and leverage this memory to assist both the training and inference processes. During training process, the gradients are constantly calibrated against the data samples from previous tasks to avoid catastrophic forgetting. 
A newly designed regularization term scaled by a loss-based parameter is used to facilitate the training of personalized models, using information from the global model. In the inference process, {\ttfamily FedMeS} further leverages the memory information to perform KNN-based Gaussian inference, further strengthening the model's personalization capability. Moreover, {\ttfamily FedMeS} exhibits a major advantage of being \emph{task-oblivious}, meaning that the inference processes for test samples from all tasks are identical, and all are expected to achieve high performance. 

We theoretically characterize the convergence of {\ttfamily FedMeS}, and empirically evaluate its performance over various dataset combinations, task constructions, task distributions, and client numbers. Over all experiments, {\ttfamily FedMeS} uniformly outperforms all baselines in terms of accuracy metrics and forgetting rate. Performance of {\ttfamily FedMeS} under different memory sizes and its execution times are also experimentally investigated.

\section{Related Work}

{\bf Personalized Federated Learning.} 
A simple idea is by deploying a global model and fine tuning parameters through gradient descent on local clients \cite{cheng2021fine,yu2020salvaging, zhang2022fine}. Meta-learning based FL methods realize model personalization through hyperparameters \cite{khodak2019adaptive}. {\ttfamily PerKNN} \cite{marfoq2022personalized} is a special case, where embeddings of training samples are stored for local memorization for KNN-based Gaussian inference. The mainstream design is to interpolate a global model and one local model per client, and the task-specific models are learned both globally and locally \cite{achituve2021personalized,shen2020federated}. Below are some methods adopting this design. Regularization terms on proximal models help construct personalized information \cite{li2021ditto,marfoq2021federated}. Sometimes an initialized model or regularizer-based learnable bases are constructed for each client \cite{huang2021personalized}. In other cases, multiple global models are trained at the server, and task-specific regularization terms force similar tasks to be close, making the training on heterogeneous clients more accurate \cite{mansour2020three,deng2020adaptive}. 

{\bf Continual Learning.} Memory replay methods are widely used in continual learning (CL) to maintain prediction accuracies of past tasks. 
Generally speaking, a memory buffer is used to store previous data which are replayed while learning a new task to alleviate forgetting \cite{wang2022memory, shim2021online}. Experience replay (ER) jointly optimizes the network parameters by interleaving the previous task exemplars with current task data \cite{riemer2018learning, isele2018selective}. An alternative solution is by constrained optimization. {\ttfamily GEM} \cite{lopez2017gradient} and {\ttfamily A-GEM} \cite{chaudhry2018efficient} leverage episodic memory to compute previous task gradients to constrain the current update step. Pesudo Rehearsal \cite{shin2017continual} is used in early works with shallow neural networks. Generative-based methods \cite{van2020brain} synthesize previous data produced by generative models and replay the synthesized data. Besides replay methods, regularization-based methods \cite{yu2020semantic,shi2021continual} and parameter isolation methods \cite{de2021continual} have also been proposed for CL.

Although a lot of work has been done in CL, just a few works have tried to use CL in a federated setting.  {\ttfamily FedWeIT} \cite{yoon2021federated} is a typical one, where task-generic and task-specific knowledge are shared across clients to decompose the model parameters. However, as demonstrated in \cite{venkatesha2022addressing}, {\ttfamily FedWeIT} struggles to handle the issue of client drift caused by data heterogeneity and does not address the scalability of tasks. Other methods, like {\ttfamily  LFedCon2} \cite{casado2020federated}, use traditional classifiers instead of DNN and propose an algorithm dealing with a concept drift based on ensemble retraining. {\ttfamily FLwF} and {\ttfamily FLwF-2T} \cite{usmanova2021distillation} use a distillation-based approach dealing with catastrophic forgetting in FL scenario and focus on the class-incremental learning scenario. 

\section{Problem Definition}\label{Problem Definition}
We consider an FL system that consists of $n$ clients and a central server. Over time, each client $k$ ($ k=1,\ldots, n$) continually collects private datasets from a sequence of $T$ machine learning tasks. For each task $t$ ($t=1,\ldots, T$), the corresponding dataset at client $k$ is denoted as $\mathcal{D}_k^t$.
We focus on a general non-IID case, where $\mathcal{D}_k^t$ is drawn from some probability distribution ${\cal P}_k^t$, and no particular relationships for ${\cal P}_k^t$ across $k$ and $t$ are assumed.

The conventional FL problem corresponds to a single-task scenario. For a particular task $t$, the following objective function is optimized over the global model $\mathbf{w}$:
\begin{equation}
\min _{\mathbf{w}} \mathcal{G}(F_{1}(\mathbf{w};\mathcal{D}_1^t), \ldots, F_{n}(\mathbf{w};\mathcal{D}_n^t))
\label{global obj} 
\end{equation}
where $F_k(\cdot)$ is a local objective for client $k$ 
and $\mathcal{G}(\cdot)$ is some aggregation function. 

In the general case of evolving tasks,
each client intends to obtain a personalized model which maintains good performance on all its previous tasks. Motivated by this need, we incorporate the concept of continual learning and into a personalized FL framework, and formally define the personalized federated continual learning (PFCL) problem below.
Specifically, for each task $t$, the personalized model $w_k^t$ for each client $k$ is obtained through
\begin{equation}
\begin{aligned}
    \min _{w_k^t} \enspace & {G}_k(w_k^t;\mathbf{w^*})= \mathcal{L}\left(w_k^t; \mathcal{D}_k^t\right)+\frac{\lambda}{2}\left\|w_k^t-\mathbf{w}^*\right\|^{2},\\
\text{s.t.} \enspace &  \mathbf{w^*} \in \underset{\mathbf{w}}{\arg \min }  \ \mathcal{G}
\left(\mathcal{L}(\mathbf{w};\mathcal{D}_1^t), ..., \mathcal{L}(\mathbf{w};\mathcal{D}_n^t)\right), \\
& \mathcal{L}\left(w_k^t; \mathbf{D}_k^{t-1}\right) \leq \mathcal{L}\left(w_k^{t-1}; \mathbf{D}_k^{t-1}\right).
\end{aligned}
\label{PFCL formal}
\end{equation}
Here $\mathbf{D}_k^{t-1} = ({\cal D}_k^1,\ldots,{\cal D}_k^{t-1})$, and $\mathcal{L}(w; \mathbf{D})$ is the empirical loss of $w$ on dataset $\mathbf{D}$. The last constraint ensures that the new model obtained from task $t$ has a lower loss than the old model over previous datasets, effectively alleviating the forgetting of previous tasks.

\begin{algorithm*}[!htb]
	\caption{{\ttfamily FedMeS}}
        \label{alg:FedMeS}
	\KwIn{Clients set $\{C_k\}_{k=1}^{n}$, each client holds its local datasets $\{\mathcal{D}_k^{t}\}_{t=1}^T$ of $T$ tasks, communication rounds $R$, test data $\mathbf{x} \in \mathcal{X}$}
	\KwOut{Local model parameter $\{w_k\}_{k=1}^{n}$, global model parameter $\mathbf{w} $, local prediction $\text{FedMeS}_k(\mathbf{x})$}  
	\BlankLine
        Initialize global parameter $\mathbf{w} $, and for each client $\mathcal{M}_k\longleftarrow \{\}$
        
        \textbf{Training process of {\ttfamily FedMeS}:}
        
	\While{\textnormal {task} $t \in [T]$ }{
            \ForEach{\textnormal {round} $r=1$ \textnormal{to} $R$}{
            Server sends $\mathbf{w} $ to all clients
            
            \ForEach{\textnormal {client} $k=1$ \textnormal {to} $n$ \textnormal{in parallel}}{
			Minimizing local PFCL problems according to \eqref{w_k update rule 1} \eqref{w_k update rule 2} to obtain the $w_k^t$, the $\lambda$ is calculated by \eqref{lam}

                Send the $w_k^t$ to the server

		}
            Server update $\mathbf{w} \gets \frac {1}{n} {\textstyle \sum_{{k=1}}^{{n}}}  w_k^t$
            }
            Each client $\mathrm{APPEND} (\mathcal{M}_k^t )= \mathrm{SAMPLE} (\mathcal{D}_k^{t})$

	}	
        \textbf{Inference Process of {\ttfamily FedMeS}:}
        
        \ForEach{\textnormal {client} $k=1$ \textnormal {to} $n$ \textnormal{in parallel}}{
			Compute the representation of all memory samples $P_{w_k}(\mathbf{m}_k^i)$ through the forward of $w_k$ to obtain the R-L pairs

                Obtain $P_{w_k}(\mathbf{x})$ and retrieve its k-nearest neighbors according to \eqref{knn calculate}, obtain the local KNN-based prediction according to \eqref{knn output}

		}
\end{algorithm*}

\section{The Proposed {\ttfamily FedMeS} Framework} 
We propose an algorithm, \textbf{F}ederated \textbf{M}emory \textbf{S}trengthening ({\ttfamily FedMeS}), to solve the PFCL problem defined in \eqref{PFCL formal}. The key idea of {\ttfamily FedMeS} is to design local memory data samples from previous tasks, and flexibly use them in both training and inference processes. In training process, the local memory is used for gradient correction to avoid catastrophic forgetting; in inference process, a KNN algorithm based on the representations of local samples helps to improve the accuracy of the personalized model. The overall workflow of {\ttfamily FedMeS} is illustrated in Figure~\ref{FedMeS_Dig}, and pseudocode is given in \cref{alg:FedMeS}. 

\subsection{Memory design}\label{Mmeory Design}
In {\ttfamily FedMeS}, we utilize a small amount of memory at each client to help meeting the constraint of $\mathcal{L}\left(w_k^t; \mathbf{D}_k^{t-1}\right) \leq \mathcal{L}\left(w_k^{t-1}; \mathbf{D}_k^{t-1}\right)$ in \eqref{PFCL formal}. Specifically, given a fixed memory size $m$, each client $k$ constructs and stores a dataset $\mathcal{M}_k^t \subset \bigcup_{i=1}^{t-1} \mathcal{D}_k^i$ with $|\mathcal{M}_k^t| \leq m$, through sub-sampling data points from each of the previous datasets.
For the initial task,  $\mathcal{M}_k^1 =\emptyset$. After completing task $t$, a small number of $m^t$ data points are randomly sampled from ${\cal D}_{k}^t$, and combined with $\mathcal{M}_k^t$ to form $\mathcal{M}_k^{t+1}$ for the next task $t+1$.
The FIFO scheduling is adopted, such that when the memory is full, data points from older tasks are popped to make space for incoming tasks. By utilizing $\mathcal{M}_k^t$ as a compact representation of $\mathbf{D}_k^{t-1}$, {\ttfamily FedMeS} aims to solve a PFCL problem, with the constraint $\mathcal{L}\left(w_k^t; \mathbf{D}_k^{t-1}\right) \leq \mathcal{L}\left(w_k^{t-1}; \mathbf{D}_k^{t-1}\right)$ in the \eqref{PFCL formal} replaced by
\begin{equation}
    \mathcal{L}\left(w_k^t; \mathcal{M}_k^t\right) \leq \mathcal{L}\left(w_k^{t-1}; \mathcal{M}_k^t\right).
    \label{change d_k to m_k}
\end{equation}

\subsection{Training Process of {\ttfamily FedMeS}}

The training of task $t$ ($t=1,\ldots,T$) proceeds over multiple global iterations between the server and the clients. In each global iteration, the server broadcasts the global model $\mathbf{w}$ to the clients, waits for clients to upload personalized models $w_k^t$, and aggregates them to update the global model $\mathbf{w} \leftarrow \frac{1}{n}\sum_{k=1}^nw_k^t$.

During local training process, each client $k$ runs multiple local iterations to update $w_k^t$. 
In each local iteration, client $k$ starts with the global model $\mathbf{w}$ and perform SGD according to the objective in \eqref{PFCL formal} to update $w_k^t$. The model update for each mini-batch is different according to whether the condition \eqref{change d_k to m_k} is satisfied. For a better compatibility with running SGD, we alternatively consider the following condition on the inner product of current gradients on current and previous tasks:
\begin{equation}
    \left\langle\nabla \mathcal{L}\left(w_k^t ; \mathcal{D}_{k}^{t}\right), \nabla \mathcal{L}\left(w_k^t ; \mathcal{M}_k^t\right)\right\rangle \geq 0,
    \label{nabla ineq}
\end{equation}
which was shown in \cite{chaudhry2018efficient} equivalent to the condition in \eqref{change d_k to m_k}.

When \eqref{nabla ineq} is satisfied, it means that the updates on current task $t$ and previous tasks are roughly in the same direction, so the optimization on current task would not negatively impact the performance of past tasks. As a result, it is safe to update the model along the gradient of current task as follows, for some learning rate $\eta_1$:

\begin{equation}
    w_k^t = w_k^t - \eta_1\left(\nabla \mathcal{L}(w_k^t; \mathcal{D}_k^t)+\lambda\Vert w_k^t-\mathbf{w}\Vert\right).
    \label{w_k update rule 1}
\end{equation}

When \eqref{nabla ineq} does not hold, client $k$ first adjusts its local weights $w_k^t$ to avoid forgetting, through the following gradient correction step, for some learning rate $\eta_2$:
\begin{equation}
\begin{aligned}
w_k^t = w_k^t-&\eta_2\bigg( \nabla \mathcal{L}(w_k^t; \mathcal{D}_k^t)
 - 
\\ &\frac{\nabla \mathcal{L}(w_k^t; \mathcal{D}_k^{t})^{\top} \nabla \mathcal{L}(w_k^t;\mathcal{M}_k^t)}{\nabla \mathcal{L}(w_k^t;\mathcal{M}_k^t)^{\top} \nabla \mathcal{L}(w_k^t;\mathcal{M}_k^t)}\nabla \mathcal{L}(w_k^t; \mathcal{M}_k^t) \bigg ).
     \label{w_k update rule 2}
\end{aligned}
\end{equation}
Note that unlike in \eqref{w_k update rule 1},  this gradient correction step occurs purely on local weight $w_k^t$ and does not involve global weights $\mathbf{w}$. The rationale for such  modifications can be found in \cref{Reasoning of Local Continual Learning}.

\begin{figure*}[!htb]
\begin{center}
\centerline{\includegraphics[width=160mm]{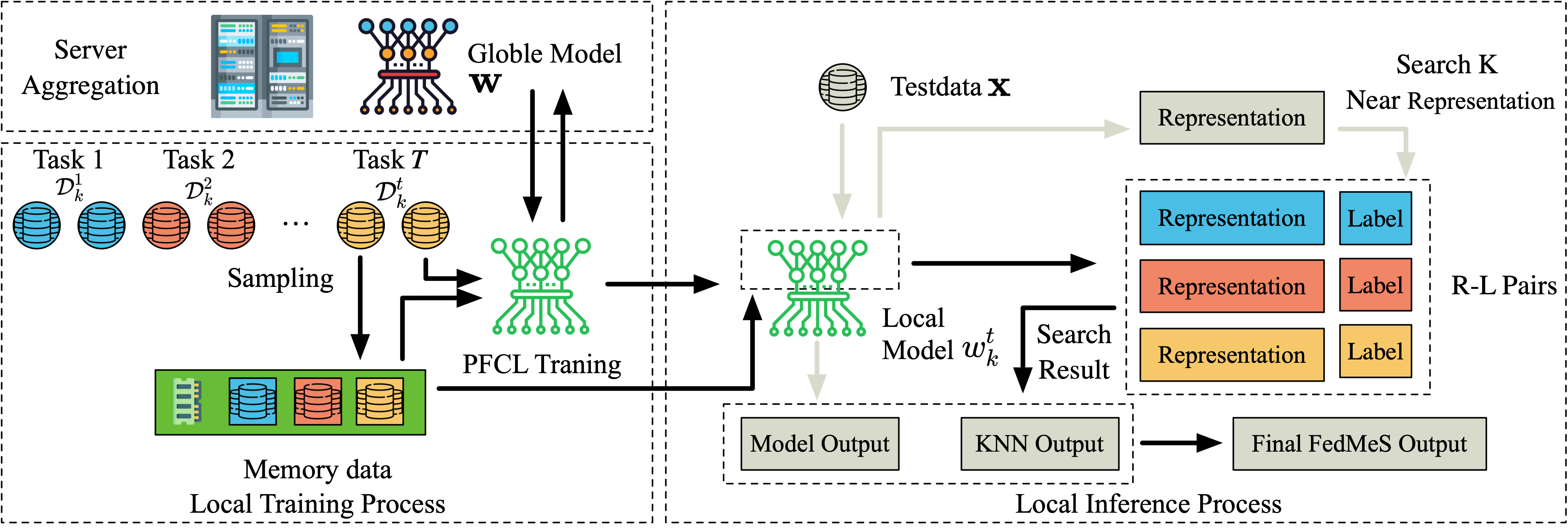}}
\vspace{-2mm}
\caption{Workflow of {\ttfamily FedMeS}. Local memory is utilized in both training and inference processes.}
\label{FedMeS_Dig}
\end{center}
\vspace{-8mm}
\end{figure*}

\begin{remark}
\textit{Only one} of \eqref{w_k update rule 1} and \eqref{w_k update rule 2} is executed in each local training iteration: gradients are updated based on the objective function in \eqref{PFCL formal} only when \eqref{nabla ineq} holds. Otherwise, as analytically demonstrated in \cref{convergence analysis}, updating $w_k^t$ according to \eqref{w_k update rule 2} multiple times allows the inner product $\langle\nabla \mathcal{L}\left(w_k^t ; \mathcal{D}_{k}^{t}\right), \nabla \mathcal{L}\left(w_k^t ; \mathcal{M}_k^t\right)\rangle$ (currently negative) to approach and eventually exceed zero.
\end{remark}

In {\ttfamily FedMeS}, rather than fixing the regularization parameter $\lambda$ in \eqref{PFCL formal}, we propose a novel loss-based approach for dynamically adjusting $\lambda$, to better draw useful information from other clients through the global model to facilitate learning of local tasks. Specifically, we set the value of $\lambda$ as:
\begin{equation}
\lambda = 2 \cdot \text{sigmoid}\left(\frac{1}{\mathcal{L}(\mathbf{w},\mathcal{D}_k^{t})}\right). 
\label{lam} 
\end{equation}
 Intuitively, when $\mathcal{L}(\mathbf{w},\mathcal{D}_k^{t})$ is relatively large, it means the global model $\mathbf{w}$ performs poorly on the current task of client $k$, and the personalized model $w_{k}^t$ should deviate from the global model by decreasing $\lambda$. On the other hand, a small $\mathcal{L}(\mathbf{w},\mathcal{D}_k^{t})$ would encourage $w_{k}^t$ to approach the global model $\mathbf{w}$, which corresponds to a lager $\lambda$. The sigmoid function here limits $\lambda$ within $[0,2]$, following the result from \cite{li2021ditto}.

 \subsection{Inference Process of {\ttfamily FedMeS}}
As shown in \cref{FedMeS_Dig}, {\ttfamily FedMeS} also utilizes local memory to enhance inference performance.
Specifically, to perform inference process after learning task $t$, client $k$ first generates a set of representation-label pairs (R-L pairs) from the current local memory as $    \left\{\left(P_{w_k^t}\left({\bf m}_k^i\right), y_{{\bf m}_k^i}\right): \left({\bf m}_k^i, y_{{\bf m}_k^i}\right) \in \mathcal{M}_k^t\right\}$, where 
${\bf m}_k^i, i=1,\ldots,|{\cal M}_k^t|$, is the input of the $i$-th sample in ${\cal M}_k^t$, and $y_{{\bf m}_k^i}$ is its label. Function $P_{w_k^t}({\bf m}_k^i)$ produces an embedding representation of ${\bf m}_k^i$. In the context of CNNs, this function can potentially generate the output of the final convolutional layer, capturing hierarchical features within the input. Likewise, in the case of transformers, it has the flexibility to produce the output of any chosen self-attention layer, facilitating comprehensive contextual encoding.

For a test sample $\mathbf{x}$ (from unknown task), we first find the $K$ nearest neighbors of $\mathbf{x}$ from the formed R-L pairs $\mathcal{K}^{(K)}(\mathbf{x})=\left\{\left(P_{w_{k}^t }({\bf m}_k^j), y_{{\bf m}_k^j}\right): 1 \leq j \leq K\right\}$ with

 \begin{equation}
dist(P_{w_{k}^t}(\mathbf{x}), P_{w_{k}^t}({\bf m}_k^j)) \leq dist(P_{w_{k}^t}(\mathbf{x}), \!P_{w_{k}^t}({\bf m}_k^{j+1})).
\label{knn calculate} 
\end{equation}
Here $dist(\cdot, \cdot)$ could be any distance metric, and in {\ttfamily FedMeS} the Euclidean distance is used. Denote  $b_{w_k^t}({\bf x})$ as the local estimate of the conditional probability $\mathcal{P}_{\mathcal{M}_k^t}(y|
\mathbf{x})$, where $\mathcal{P}_{\mathcal{M}_k^t}$ is the probability distribution of $\mathcal{M}_k^t$. 
Then the $K$ nearest neighbours found in $\mathcal{K}^{(K)}(\mathbf{x})$ are used to compute $[b_{w_k^t}(\mathbf{x})]_y$ with a Gaussian kernel:
\begin{equation}
        [b_{w_k^t}(\mathbf{x})]_y \!\propto\!\! \sum_{j=1}^{K} \mathbf{1}_{y=y_{{\bf m}_k^j}}{\!\times\!} \exp \{-dist(P_{w_{k}^t}\!(\mathbf{x}),P_{w_{k}^t}\!({\bf m}_k^j))\}.
\end{equation}
Finally, the {\ttfamily FedMeS} prediction result of $\mathbf{x}$ on client $k$ is obtain by a convex combination of outputs from local personalized model and KNN inference:
\begin{equation}
    \mathrm{FedMeS}_k(\mathbf{x}) = \theta_k \cdot b_{w_k^t}(\mathbf{x}) +(1-\theta_k)h_{w_k^t}(\mathbf{x}).
\label{knn output} 
\end{equation}
Here $h_{w_k^t}$ is the personalized local model parameterized by $w_k^t$, 
and $\theta_k \in (0, 1) $ is a combining parameter which can be tuned through local validation or cross-validation. 

\begin{remark}
As proved in \cite{khandelwal2019generalization}, augmenting the model inference with a memorization mechanism (KNN in this case) helps to improve the performance. In \cite{marfoq2022personalized}, local memorization through KNN has been applied to improve the accuracies of local models in personalized FL, for a single task. {\ttfamily FedMeS} extends the application of this technique on memorization over \emph{an arbitrary sequence of tasks}, via utilizing a subset of samples from each task. Also, this inference enhancement of {\ttfamily FedMeS} comes \emph{for free}, as this memory is readily available from the preceeding training process. 
\end{remark}
\begin{remark}
Another major advantage of {\ttfamily FedMeS} is that it is \emph{task-oblivious}. That is, the same inference process is applied for \emph{all} test samples, and no prior knowledge is needed about which task the sample belongs to.  
This also reflects the strong robustness of {\ttfamily FedMeS}: regardless of the original task, a good inference performance is always guaranteed by a unified {\ttfamily FedMeS} inference process. This is, however, not the case for other task-incremental learning CL methods like in \cite{delange2021continual}. 
\end{remark}

\subsection{Convergence Analysis}
\label{convergence analysis}
We analyze the convergence performance of {\ttfamily FedMeS} in Theorem \ref{theroem converge to *} and \ref{theorem ditto local convergence-mainbody}, whose detailed proofs are given in \cref{Convergence Analysis of the Local Training Process} and \cref{Proof of Theorem 2} respectively. 

We first make some assumptions to facilitate the analysis. For each global round $r$ of {\ttfamily FedMeS} to solve task $t$, we denote $w_k^{(r)}, \mathbf{w}^{(r)}$ respectively as the value of $w_k^t$ and $\mathbf{w}$ at round $r$. 

\vspace{-1mm}
\begin{assumption} The loss function $\mathcal{L}(w_k)$ is $c$-strongly convex and $L$-smooth for $k=1, ..., n$. The expectation of stochastic gradients of the loss function $\mathcal{L}(w_k^{(r)})$ on $\mathcal{D}_k^t$ is uniformly bounded at all clients and all iterations, i.e.: $\mathbb{E}[\Vert\nabla\mathcal{L}\left(w_k^{(r)}; \xi_k^r\right)\Vert^2] \leq \sigma^2$, where $\xi_k^r$ is the mini-batch data sampled from $\mathcal{D}_k^t$ in round $r$. The global model converges with rate $g(r)$, i.e., there exists $g(r)$ such that $\text{lim}_{r\rightarrow \infty}g(r) = 0$, $\mathbb{E}[\Vert \mathbf{w}^{(r)}-\mathbf{w}^*\Vert^2] \leq g(r)$.
\label{assumption 1-3}
\end{assumption}
\vspace{-2mm}

When \eqref{nabla ineq} is not satisfied, {\ttfamily FedMeS} executes \eqref{w_k update rule 2} to perform gradient correction. We denote the iteration index of repeating \eqref{w_k update rule 2} as $s (s=1,2,...)$, and $\zeta_k^s$ as the mini-batch data sampled from $\mathcal{M}_k^s$ in iteration $s$, and make the following two assumptions.

\begin{assumption}There exists scalars $\mu_G\geq\mu>0$ such that for all $s \in \mathbb{N}$,
     \begin{equation}
    \begin{aligned}
        \nabla\mathcal{L}\left(w_k^{(s)};  \mathcal{M}_k^t\right)^\top\mathbb{E}_{\zeta_k^s}[\nabla\mathcal{L}\left(w_k^{(s)}; \zeta_k^s\right)] \geq\\ 
 \mu\Vert\nabla\mathcal{L}&\left(w_k^{(s)}; \mathcal{M}_k^t\right)\Vert^2_2,\\
\Vert\mathbb{E}_{\zeta_k^s}[\nabla\mathcal{L}\left(w_k^{(s)}; \zeta_k^s\right)]\Vert_2 \leq \mu_G\Vert\nabla\mathcal{L}&\left(w_k^{(s)}; \mathcal{M}_k^t\right)\Vert_2.
    \end{aligned}
    \end{equation}
\label{assumption 4}
\end{assumption}
\vspace{-3mm}
\begin{assumption}There exists scalars $M\geq 0$ and $M_V\geq 0$ such that, for all $s\in \mathbb{N}$,
    \begin{equation}
        \mathbb{V}_{\zeta_k^s}[\nabla\mathcal{L}(w_k^{(s)}; \zeta_k^s)] \leq M+M_V\Vert\nabla\mathcal{L}(w_k^{(s)}; \mathcal{M}_k^t)\Vert^2_2.
    \end{equation}
\label{assumption 5}
\vspace{-2mm}
\end{assumption}
\vspace{-4mm}
\begin{theorem}
    Under Assumption \ref{assumption 1-3}, \ref{assumption 4} and \ref{assumption 5}, and the condition that the constraint \eqref{nabla ineq} is not satisfied, define $w_k^* = \mathop{\arg\min}_{w_k} \mathcal{L}(w_k; \mathcal{M}_k^t)$, and we have for $s \geq 2$,
    \vspace{-1mm}
    \begin{equation}
    \mathbb{E}[\mathcal{L}(w_k^{(s)}; \mathcal{M}_k^t) - \mathcal{L}(w_k^*; \mathcal{M}_k^t)] \leq \frac{LM}{2c^2\mu^2}.
    \end{equation}
    \vspace{-3mm}
    \label{theroem converge to *}
\end{theorem}
\vspace{-3mm}
Theorem \ref{theroem converge to *} shows that, executing the gradient correction step iteratively ensures that $\mathcal{L}(w_k^t; \mathcal{M}_k^t)$ converges to its local optimum. Therefore, $\mathcal{L}\left(w_k^t; \mathcal{M}_k^t\right)$ would eventually be less than $\mathcal{L}\left(w_k^{t-1}; \mathcal{M}_k^t\right)$ (which can be viewed as a constant that is larger than $\mathcal{L}(w_k^*; \mathcal{M}_k^t)$) with high possibility, satisfying the constraint \eqref{change d_k to m_k} (and equivalently constraint \eqref{nabla ineq} ) again. 

\begin{theorem}
Under Assumption \ref{assumption 1-3}, and the condition that the constraint \eqref{nabla ineq} is satisfied, there exists a constant $C$ such that for $\lambda \in \mathbb{R}$, $w_k^r$ converges to $w_k^* := \mathop{\arg\min}_{w_k^t} G_k\left(w_k^t; \mathbf{w}^*\right)$ with rate $Cg(r)$.
\label{theorem ditto local convergence-mainbody}
\end{theorem}
\vspace{-3mm}
By Theorem \ref{theorem ditto local convergence-mainbody}, the local model $w_k^t$ would always converge to its optimum with a multiplicative gap from the rate of global convergence process.

\begin{table*}[!htb]
\caption{$Acc\_ALL$(Acc) and Average Forgetting Rate (FR) over All Clients and All Learned Tasks.}

\begin{center}
\begin{small}
\scalebox{1}{
\begin{tabular}{lllllllll}
\hline\hline\hline
Datasets                             & \multicolumn{4}{c}{Split CIFAR-100}                                                                                                                                                 & \multicolumn{4}{c}{Split EMNIST}                                                                                                                                \\ \hline
\multicolumn{1}{l|}{client number}   & \multicolumn{2}{c|}{10}                                                                  & \multicolumn{2}{c|}{20}                                                                  & \multicolumn{2}{c|}{10}                                                                  & \multicolumn{2}{c}{20}                                               \\ \hline
\multicolumn{1}{l|}{methods}         & \multicolumn{1}{c|}{Acc}                    & \multicolumn{1}{c|}{FR}                    & \multicolumn{1}{c|}{Acc}                    & \multicolumn{1}{c|}{FR}                    & \multicolumn{1}{c|}{Acc}                    & \multicolumn{1}{c|}{FR}                    & \multicolumn{1}{c|}{Acc}                    & \multicolumn{1}{c}{FR} \\ \hline
\multicolumn{1}{l|}{FedAvg}          & \multicolumn{1}{l|}{.249 $\pm$.02}          & \multicolumn{1}{l|}{.35 $\pm$.05}          & \multicolumn{1}{l|}{.263 $\pm$.03}          & \multicolumn{1}{l|}{.29 $\pm$.03}          & \multicolumn{1}{l|}{.450 $\pm$.02}          & \multicolumn{1}{l|}{.49 $\pm$.02}          & \multicolumn{1}{l|}{.465 $\pm$.03}          & .48 $\pm$.02           \\
\multicolumn{1}{l|}{Ditto}           & \multicolumn{1}{l|}{.219 $\pm$.02}          & \multicolumn{1}{l|}{.37 $\pm$.03}          & \multicolumn{1}{l|}{.221 $\pm$.02}          & \multicolumn{1}{l|}{.38 $\pm$.06}          & \multicolumn{1}{l|}{.388 $\pm$.03}          & \multicolumn{1}{l|}{.71 $\pm$.04}          & \multicolumn{1}{l|}{.381 $\pm$.02}          & .72 $\pm$.03           \\
\multicolumn{1}{l|}{FedRep}          & \multicolumn{1}{l|}{.415 $\pm$.04}          & \multicolumn{1}{l|}{.13 $\pm$.02}          & \multicolumn{1}{l|}{.425 $\pm$.06}          & \multicolumn{1}{l|}{.13 $\pm$.02}          & \multicolumn{1}{l|}{.723 $\pm$.04}          & \multicolumn{1}{l|}{.23 $\pm$.04}          & \multicolumn{1}{l|}{.723 $\pm$.06}          & .24 $\pm$.06           \\
\multicolumn{1}{l|}{FedAGEM}         & \multicolumn{1}{l|}{.351 $\pm$.04}          & \multicolumn{1}{l|}{.14 $\pm$.03}          & \multicolumn{1}{l|}{.398 $\pm$.05}          & \multicolumn{1}{l|}{.14 $\pm$.03}          & \multicolumn{1}{l|}{.817 $\pm$.04}          & \multicolumn{1}{l|}{.09 $\pm$.05}          & \multicolumn{1}{l|}{.828 $\pm$.04}          & .16 $\pm$.04           \\
\multicolumn{1}{l|}{FedWEIT}         & \multicolumn{1}{l|}{.421 $\pm$.05}          & \multicolumn{1}{l|}{.06 $\pm$.03}          & \multicolumn{1}{l|}{.432 $\pm$.05}          & \multicolumn{1}{l|}{.08 $\pm$.03}          & \multicolumn{1}{l|}{.867 $\pm$.03}          & \multicolumn{1}{l|}{.03 $\pm$.01}          & \multicolumn{1}{l|}{.857 $\pm$.03}          & .03 $\pm$.02           \\
\multicolumn{1}{l|}{FedMeS-noLIP}    & \multicolumn{1}{l|}{.475 $\pm$.05}          & \multicolumn{1}{l|}{.08 $\pm$.03}          & \multicolumn{1}{l|}{.488 $\pm$.05}          & \multicolumn{1}{l|}{.07 $\pm$.03}          & \multicolumn{1}{l|}{.891 $\pm$.02}          & \multicolumn{1}{l|}{.03 $\pm$.05}          & \multicolumn{1}{l|}{.912 $\pm$.02}          & .02 $\pm$.05           \\
\multicolumn{1}{l|}{\textbf{FedMeS}} & \multicolumn{1}{l|}{\textbf{.530 $\pm$.05}} & \multicolumn{1}{l|}{\textbf{.06 $\pm$.01}} & \multicolumn{1}{l|}{\textbf{.533 $\pm$.04}} & \multicolumn{1}{l|}{\textbf{.06 $\pm$.02}} & \multicolumn{1}{l|}{\textbf{.935 $\pm$.01}} & \multicolumn{1}{l|}{\textbf{.01 $\pm$.01}} & \multicolumn{1}{l|}{\textbf{.964 $\pm$.01}} & \textbf{.01 $\pm$.01}  \\ \hline
\multicolumn{1}{l|}{datasets}        & \multicolumn{4}{c|}{Split CORe50}                                                                                                                                                   & \multicolumn{4}{c}{Split MiniImageNet}                                                                                                                          \\ \hline
\multicolumn{1}{l|}{client number}   & \multicolumn{2}{c|}{10}                                                                  & \multicolumn{2}{c|}{20}                                                                  & \multicolumn{2}{c|}{10}                                                                  & \multicolumn{2}{c}{20}                                               \\ \hline
\multicolumn{1}{l|}{methods}         & \multicolumn{1}{c|}{Acc}                    & \multicolumn{1}{c|}{FR}                    & \multicolumn{1}{c|}{Acc}                    & \multicolumn{1}{c|}{FR}                    & \multicolumn{1}{c|}{Acc}                    & \multicolumn{1}{c|}{FR}                    & \multicolumn{1}{c|}{Acc}                    & \multicolumn{1}{c}{FR} \\ \hline
\multicolumn{1}{l|}{FedAvg}          & \multicolumn{1}{l|}{.303 $\pm$.01}          & \multicolumn{1}{l|}{.60 $\pm$.02}          & \multicolumn{1}{l|}{.311 $\pm$.01}          & \multicolumn{1}{l|}{.59 $\pm$.02}          & \multicolumn{1}{l|}{.271 $\pm$.02}          & \multicolumn{1}{l|}{.51 $\pm$.04}          & \multicolumn{1}{l|}{.262 $\pm$.02}          & .39 $\pm$.05           \\
\multicolumn{1}{l|}{Ditto}           & \multicolumn{1}{l|}{.266 $\pm$.01}          & \multicolumn{1}{l|}{.79 $\pm$.02}          & \multicolumn{1}{l|}{.267 $\pm$.01}          & \multicolumn{1}{l|}{.81 $\pm$.03}          & \multicolumn{1}{l|}{.264 $\pm$.03}          & \multicolumn{1}{l|}{.49 $\pm$.05}          & \multicolumn{1}{l|}{.265 $\pm$.01}          & .51 $\pm$.04           \\
\multicolumn{1}{l|}{FedRep}          & \multicolumn{1}{l|}{.547 $\pm$.03}          & \multicolumn{1}{l|}{.34 $\pm$.02}          & \multicolumn{1}{l|}{.551 $\pm$.04}          & \multicolumn{1}{l|}{.35 $\pm$.03}          & \multicolumn{1}{l|}{.410 $\pm$.03}          & \multicolumn{1}{l|}{.30 $\pm$.05}          & \multicolumn{1}{l|}{.388 $\pm$.03}          & .24 $\pm$.02           \\
\multicolumn{1}{l|}{FedAGEM}         & \multicolumn{1}{l|}{.731 $\pm$.04}          & \multicolumn{1}{l|}{.18 $\pm$.03}          & \multicolumn{1}{l|}{.741 $\pm$.04}          & \multicolumn{1}{l|}{.20 $\pm$.03}          & \multicolumn{1}{l|}{.504 $\pm$.05}          & \multicolumn{1}{l|}{.18 $\pm$.03}          & \multicolumn{1}{l|}{.477 $\pm$.05}          & .17 $\pm$.05           \\
\multicolumn{1}{l|}{FedWEIT}         & \multicolumn{1}{l|}{.595 $\pm$.04}          & \multicolumn{1}{l|}{.17 $\pm$.04}          & \multicolumn{1}{l|}{.589 $\pm$.05}          & \multicolumn{1}{l|}{.19 $\pm$.04}          & \multicolumn{1}{l|}{.319 $\pm$.04}          & \multicolumn{1}{l|}{.17 $\pm$.03}          & \multicolumn{1}{l|}{.343 $\pm$.04}          & .15 $\pm$.03           \\
\multicolumn{1}{l|}{FedMeS-noLIP}    & \multicolumn{1}{l|}{.821 $\pm$.04}          & \multicolumn{1}{l|}{.05 $\pm$.03}          & \multicolumn{1}{l|}{.834 $\pm$.04}          & \multicolumn{1}{l|}{.04 $\pm$.03}          & \multicolumn{1}{l|}{.607$\pm$.04}           & \multicolumn{1}{l|}{.11 $\pm$.03}          & \multicolumn{1}{l|}{.617$\pm$.04}           & .11 $\pm$.03           \\
\multicolumn{1}{l|}{\textbf{FedMeS}} & \multicolumn{1}{l|}{\textbf{.877 $\pm$.04}} & \multicolumn{1}{l|}{\textbf{.04 $\pm$.01}} & \multicolumn{1}{l|}{\textbf{.891 $\pm$.04}} & \multicolumn{1}{l|}{\textbf{.03 $\pm$.01}} & \multicolumn{1}{l|}{\textbf{.645 $\pm$.05}} & \multicolumn{1}{l|}{\textbf{.08 $\pm$.02}} & \multicolumn{1}{l|}{\textbf{.659 $\pm$.03}} & \textbf{.08 $\pm$.02}  \\ \hline\hline\hline
\end{tabular}%
}
\end{small}
\label{acc and FR 1}
\end{center}

\end{table*}

\section{Experiments}
\subsection{Setup}
\textbf{Datasets and models.}
We select five commonly used public datasets: CIFAR-100 \cite{krizhevsky2009learning}, EMINIST \cite{cohen2017emnist}, CORe50 \cite{lomonaco2017core50}, MiniImageNet-100 \cite{vinyals2016matching} and TinyImageNet-200. For the purpose of PFCL evaluation, we split these datasets into multiple tasks forming four \textit{cross-class datasets}: \textbf{Split CIFAR-100}: CIFAR-100 consists of 100 classes, we split them into 10 tasks with 10 classes each.
\textbf{Split EMINIST}: We utilize 60 of the 62 categories in the original dataset, and split them into 10 tasks with 6 classes each. A total of 120,000 imagines are used.
\textbf{Split CORe50}: CORe50 is specifically designed for assessing continual learning techniques and has 50 objects collected in 11 different sessions. We naturally split it into 11 tasks with 50 classes each. 
\textbf{Split MiniImageNet}: MiniImageNet-100 is commonly used in few-shot learning benchmarks, which consists of 50,000 data points and 10,000 testing points from 100 classes. We split this dataset into 10 tasks with 10 classes each.

Besides, we also design \textit{cross-domain datasets} to evaluate the cross domain performance for {\ttfamily FedMeS}. \textbf{Fusion Tasks-40}: This benchmark combines 200,000 images from three distinct datasets: CIFAR-100, MiniImageNet-100, and TinyImageNet-200, with a total of 400 classes. These classes are then divided into 40 non-IID tasks, with each task comprising 10 disjoint classes. 

We use 6-layer CNNs for the Split CIFAR-100 and Split CORe50, 2-layer CNNs for the Split EMINIST, and ResNet-18 for Split MiniImageNet and Fusion Tasks-40. 
For the task and dataset distributions, each client is assigned a unique task sequence, in which a subset of 2-5 classes are randomly selected for each task, with the goal of ensuring data heterogeneity. 

\textbf{Metrics.} We consider the following performance metrics. 

$\bullet$ \textbf{Average Accuracy:} We consider four different average accuracies. We define the averaged accuracy of client $k$ among all learned $t$ tasks after the training of task $t$: $A_{k,t} = \frac{1}{t}  {\textstyle \sum_{i=1}^{t}} a_{t,i}^{k}$ as accuracy of client $k$ at task $t$, where $a_{t,i}^{k}$ ($i<t$) is the test accuracy of task $i$ after the training of task $t$ in client $k$; averaged accuracy of client $k$ after training all $T$ tasks  $Acc\_Client_{k} = \frac{1}{T}  {\textstyle \sum_{i=1}^{T}} A_{k,i}$; average accuracy among all $n$ clients at $t$-th task: $Acc\_Task_{t} = \frac{1}{n}  {\textstyle \sum_{j=1}^{n}} A_{j,t}$; average accuracy among all $n$ clients in all learned $T$ tasks after completing the training process of all tasks: $Acc\_ALL = \frac{1}{n}\frac{1}{T}  {\textstyle \sum_{j=1}^{n}}{\textstyle \sum_{i=1}^{T}} A_{j,t}$.

$\bullet$ \textbf{Forgetting rate:} The forgetting rate $F_t^k$ of client $k$ at $t$-th task is defined as $F_t^k=\frac{1}{t-1}  {\textstyle \sum_{i=1}^{t-1}} \max_{j\in \{1,\dots, t-1 \}}(a_{j,i}^k-a_{t,i}^k)$, which measures the maximum accuracy loss when incorporating new tasks. Forgetting rate indicates the capability of resisting catastrophic forgetting.

\begin{figure*}[!htb]
\centerline{\includegraphics[width=170mm]{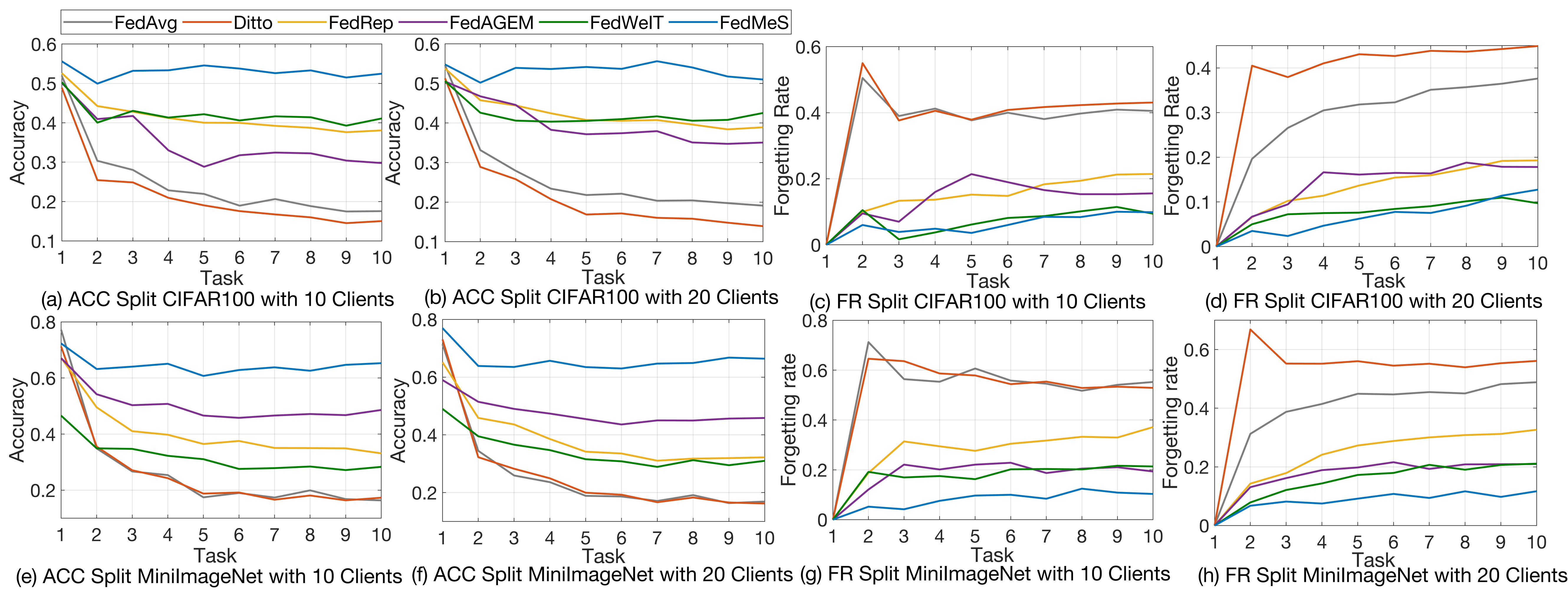}}
\caption{(a - d) Average accuracy and forgetting rate among all clients over 10 tasks on Split CIFAR-100 with 10 and 20 clients; (e - h) Average accuracy and forgetting rate among all clients over 10 tasks on Split MiniImageNet with 10 and 20 clients.}
\label{Average Acc and Forgetting Rate cifar 100}
\end{figure*}

\begin{figure*}[!htb]
\centerline{\includegraphics[width=170mm]{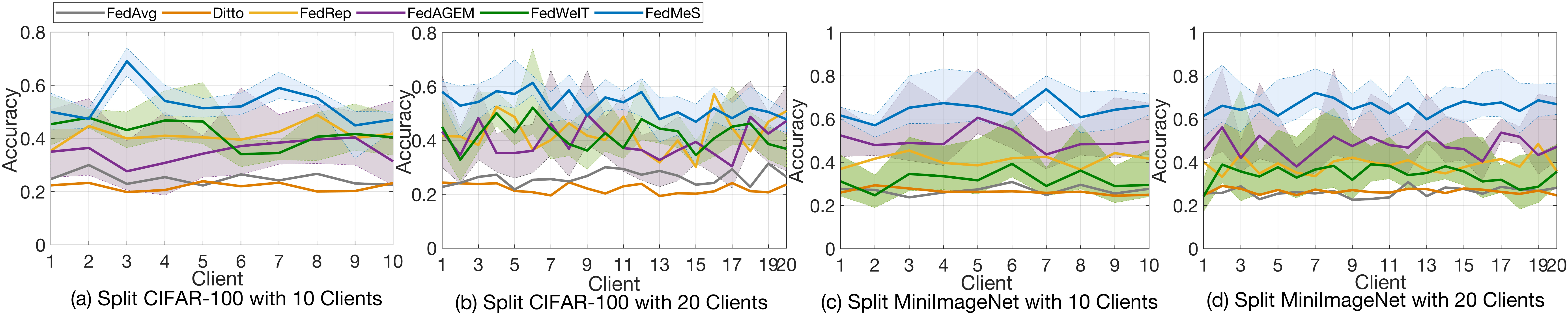}}
\caption{(a) and (b) Client accuracies averaged over all learned tasks on Split CIFAR-100 with 10 and 20 clients; (c) and (d) Client accuracies averaged over all learned tasks on Split MiniImageNet with 10 and 20 clients. The shaded area is the accuracy range of tasks for each client.}
\label{Average Accuracy in all learned tasks}
\end{figure*}

\textbf{Baselines.} 
We compare our proposed {\ttfamily FedMeS} with other personalized FL and FCL baselines. 
$\bullet$ {\ttfamily FedAvg}: A classical FL method which the server aggregates the models for all clients according to a weighted averaging of model parameters in each clients.

$\bullet$ {\ttfamily Ditto}: A simple personalized FL method that utilizes a regularization term addressing the accuracy, robustness and fairness in FL while optimizing communication efficiency.  

$\bullet$ {\ttfamily FedRep}: A personalized FL method that learns a divided model with global representation and personalized heads. Only the global representation is communicated between the server and clients, while each client adapts to its personalized head locally. 

$\bullet$ {\ttfamily FedAGEM}: An FCL method that combines the conventional {\ttfamily A-GEM} method with {\ttfamily FedAvg}. 

$\bullet$ {\ttfamily FedWeIT}: State-of-the-art FCL approach based on parameter isolation, which uses masks to divide the model parameter into base and task-adaptive parameters. The server averages the base parameters and broadcasts the task-adaptive parameters to all clients. Each client then trains all the task-adaptive parameters with the new task's weights parameters based on a regularized objective. 

\textbf{Descriptions of Hyperparameters.} For all the experiment, Adam optimizer is applied with adaptive learning rate which decay per round. Local sample batch size is 40. The participation rate(fraction) for the clients is set to 1. For Split CIFAR-100, Split EMINIST, Split CORe50 and Split MiniImageNet, 10 local epochs is performed in each communication round. And in Fusion Tasks-40, 5 local epochs is performed in each communication round. For each task a total of 10 communication rounds is performed. The regularization parameter $\lambda$ in {\ttfamily Ditto} is set to 0.1. Number of memory samples in {\ttfamily FedAGEM} and proposed {\ttfamily FedMeS} is set to 150. The $\theta_{k}$ in {\ttfamily FedMeS} is set to 0.5, and the number of nearest neighbours is set to 9.

All the experiments are conducted using PyTorch version 1.9 on a single machine equipped with two Intel Xeon 6226R CPUs, 384GB of memory, and four NVIDIA 3090 GPUs. 
Each experiment is repeated for 5 times. The averages and standard deviations of the above metrics are reported.

\begin{figure*}[!htb]
\centerline{\includegraphics[width=170mm]{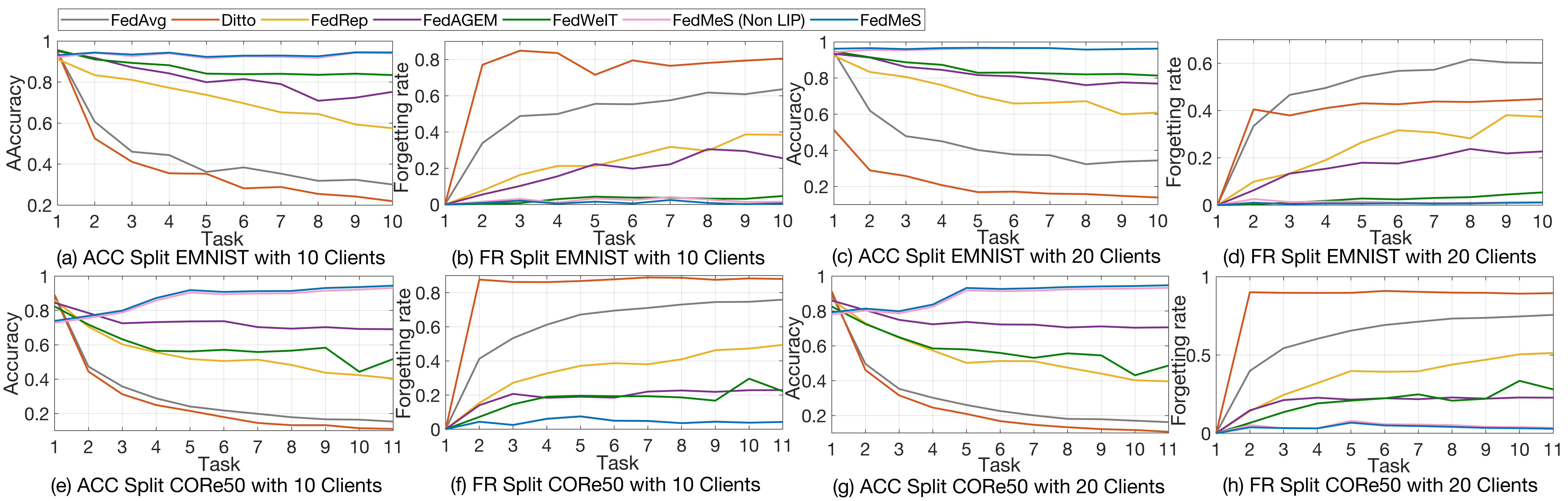}}
\caption{(a - d) Average accuracy and Average forgetting rate among all clients in all learned tasks
at x-th task on Split EMINIST with 10 clients and 20 clients. (e - h) Average accuracy and Average
forgetting rate among all clients in all learned tasks at x-th task on Split CORe50 with 10 clients
and 20 clients.}
\label{Average Acc and Forgetting Rate emnist}
\end{figure*}

\begin{figure*}[!htb]
\centerline{\includegraphics[width=170mm]{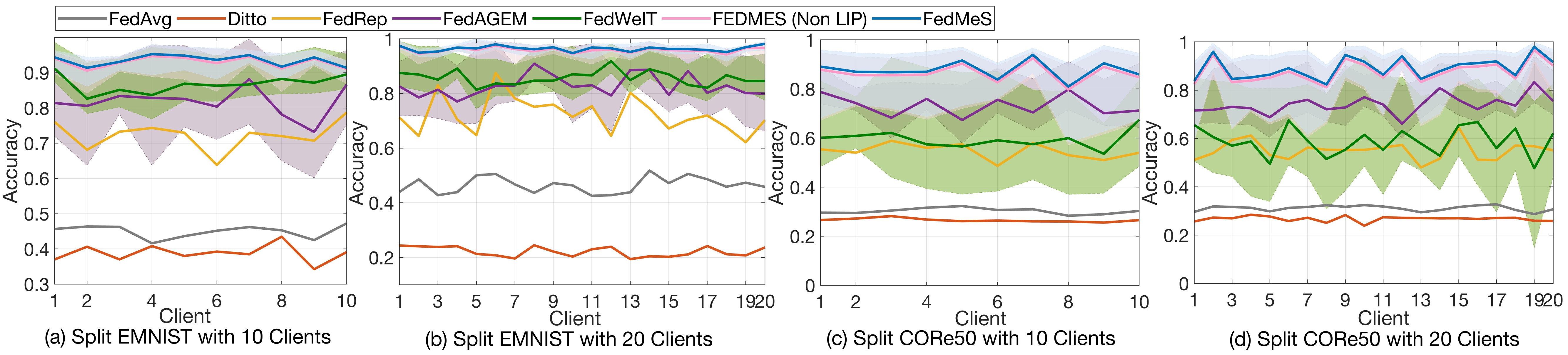}}
\caption{(a) and (b) Average accuracy of client $x$ among all learned tasks on Split EMINIST with 10 clients and 20 clients. (c) and (d) Average accuracy of client $x$ among all learned tasks on  Split CORe50 with 10 clients and 20 clients. The shade area is the accuracy range of tasks in each client.}
\label{Average Accuracy in all learned tasks emnist core}
\vskip -0.1in
\end{figure*}

\subsection{Cross-class Performance}
We observe in \cref{acc and FR 1} that, for all cross-class datasets and number of clients, {\ttfamily FedMeS} outperforms all baselines in terms of average accuracy and forgetting rate. Notably, when compared to the state-of-the-art baseline {\ttfamily FedWeIT}, {\ttfamily FedMeS} achieves a performance boost of up to $102\%$ in Split MiniImageNet. Additionally, {\ttfamily FedMeS} consistently outperforms {\ttfamily FedWeIT} in terms of forgetting rate, achieving a reduction of 4 to 6 times on the Split CORe50 dataset.
Here {\ttfamily FedWeIT} experiences a significant decline in performance when applied to the Split MiniImageNet with ResNet-18. This is primarily due to its requirement of modifying the model structure to decompose the model parameters individually. Specifically, the downsample layers in ResNet-18 contain a relatively small number of essential parameters, and decomposing these layers negatively impacted the model's accuracy. Additionally, the local inference process used in {\ttfamily FedMeS} has shown promising improvements (5$\%$ to 10$\%$ better performance in $Acc\_ALL$(Acc)), particularly when compared to {\ttfamily FedMeS} (noLIP).

\textbf{Catastrophic forgetting.} As shown in \cref{Average Acc and Forgetting Rate cifar 100}, {\ttfamily FedMeS} consistently achieves the highest accuracy and lowest forgetting rate, for all tasks under all settings. Catastrophic forgetting causes serious limitation for
{\ttfamily FedAvg} and {\ttfamily Ditto}, as they do not incorporate previous task information in training. {\ttfamily FedRep} exhibits certain robustness against heterogeneity over clients. 
But without a designed mechanism to address catastrophic forgetting, it still subjects to gradual decay in average accuracy as new tasks arrive. The isolation method of {\ttfamily FedWeIT} to obtain adaptive weights on the clients cannot well maintain the knowledge from the previous tasks, resulting in a lower accuracy than {\ttfamily FedMeS}. 

\begin{figure*}[!htb]
\centerline{\includegraphics[width=170mm]{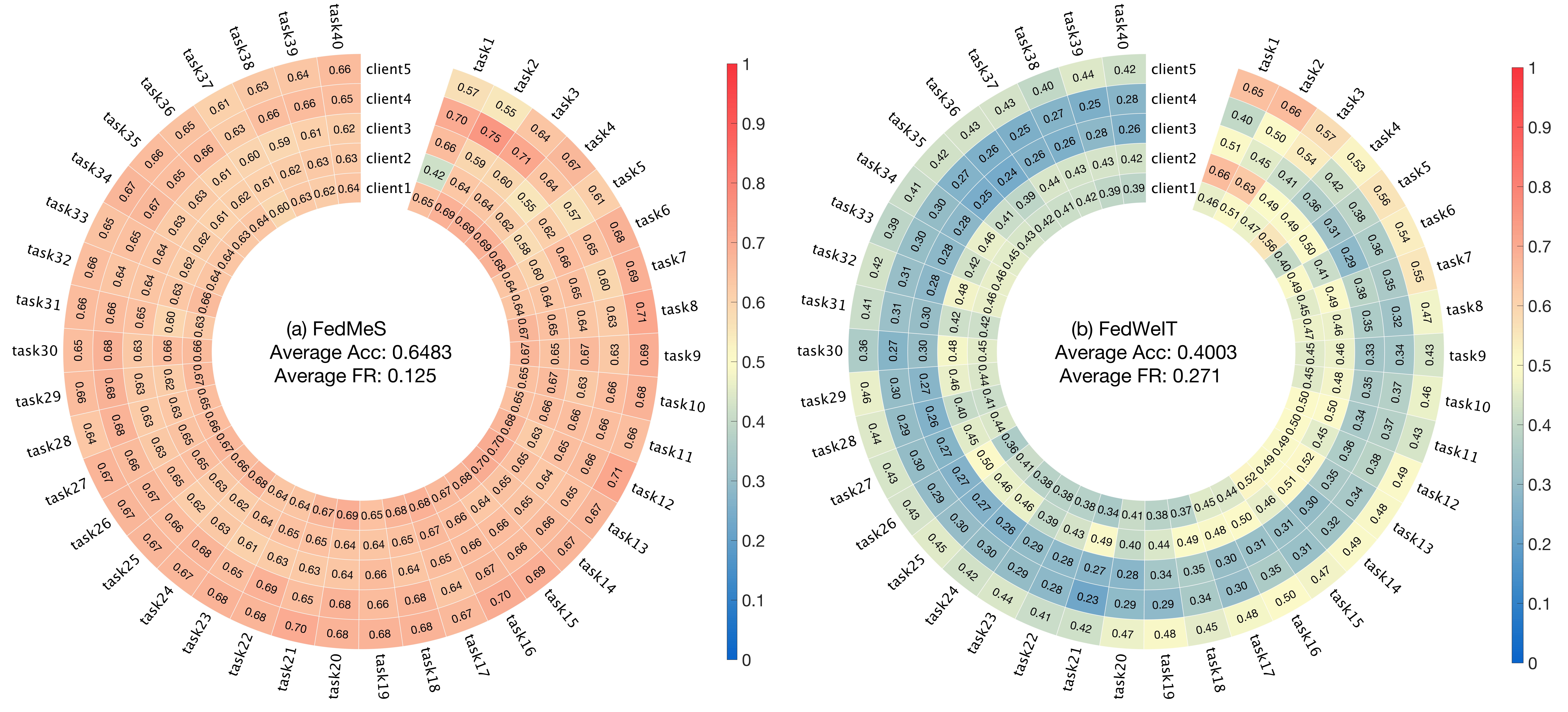}}
\caption{Averaged test accuracy of each client among all learned $t$ tasks after the training of task $t$ on Fusion Tasks-40.}
\label{task40_ALL_conpare}
\end{figure*}
\begin{figure*}[!htb]
\centerline{\includegraphics[width=170mm]{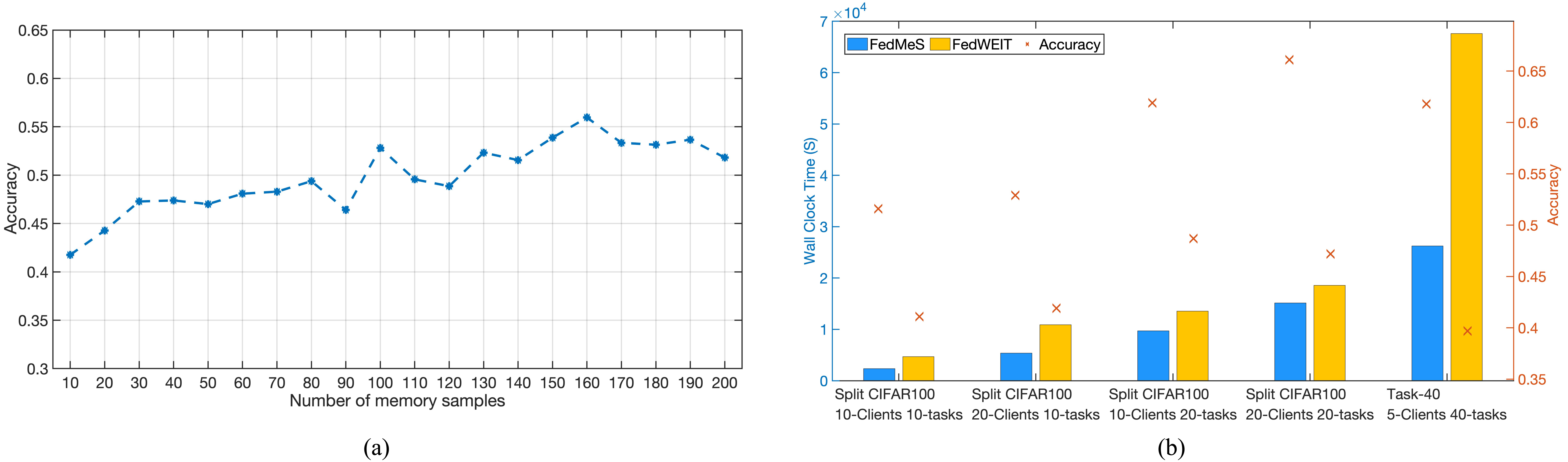}}
\caption{(a) $Acc\_ALL$ of {\ttfamily FedMeS} on Split CIFAR-100 with different number of memory samples for each client (total 10 clients). (b) Wall Clock Time and accuracy performance in different experiment settings.}
\label{Time and ACC}
\vspace{-3mm}
\end{figure*}

\textbf{Client drift.} As shown in \cref{Average Accuracy in all learned tasks}, in all settings, {\ttfamily FedMeS} achieves the highest accuracy over all clients, thanks to its reliable personalization mechanism and local inference process. 
Also, {\ttfamily FedMeS} has the narrowest shaded area over all clients, indicating its ability to maintain consistent performance across all tasks for all clients. {\ttfamily FedAGEM} and {\ttfamily FedWeIT} fail to effectively address data heterogeneity, resulting in inferior model performance. 
{\ttfamily FedWeIT} relies on the stored knowledge of all tasks at the server, which may dilute the impact of individual tasks of each client. 

 \cref{Average Acc and Forgetting Rate emnist} to \cref{Average Accuracy in all learned tasks emnist core} present the $Acc\_Task$ and $Acc\_Client$ for the Split EMINIST and Split CORe50 datasets. For every experiments {\ttfamily FedMeS} achieves highest $Acc\_Task$ and lowest forgetting rate, and for each client {\ttfamily FedMeS} achieves the best $Acc\_Client$. 

\subsection{Cross-domain Performance}
\cref{task40_ALL_conpare} shows the averaged test accuracy of each client across all tasks in the Fusion Tasks-40 dataset. The results indicate that the proposed {\ttfamily FedMeS} method outperforms the {\ttfamily FedWeIT} method, with higher average accuracy and lower forgetting rate among all clients and tasks. The poor performance of certain clients has a significant impact on other clients, and the isolation method employed by {\ttfamily FedWeIT} to mitigate catastrophic forgetting proves to be ineffective in the considered dataset with a large number of tasks. In contrast, the proposed {\ttfamily FedMeS} method achieves consistent performance across all tasks and clients, demonstrating its capability in scaling to large PFCL systems.

\begin{figure}[H]
\begin{center}
\centerline{\includegraphics[width=8.2cm]{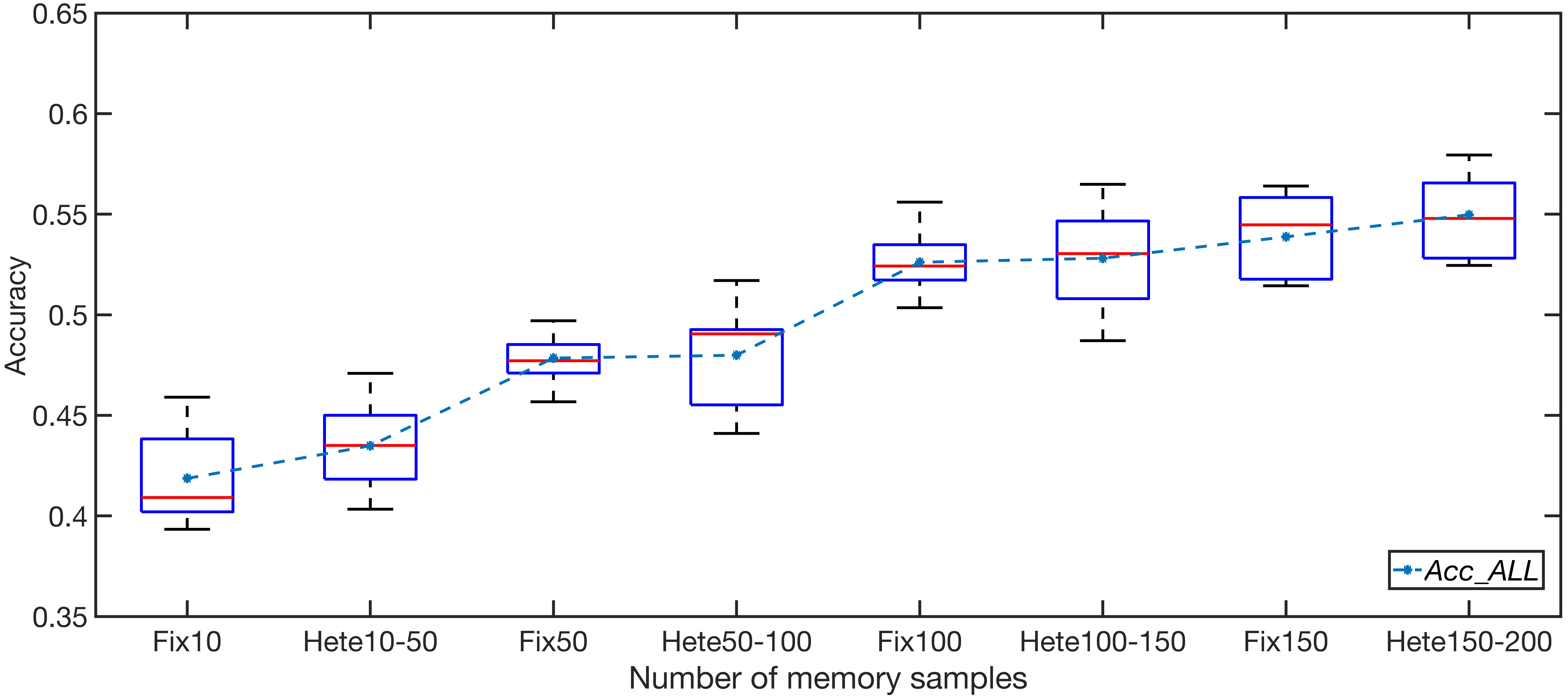}}
\caption{$Acc\_ALL$ plot and $Acc\_Task$ boxplot of {\ttfamily FedMeS} on Split CIFAR-100 with heterogeneous memory sizes for each client (total 10 clients).}
\label{mem here}
\end{center}
\vspace{-8mm}
\end{figure}

{\bf Effect of memory size.} We observe in \cref{Time and ACC} (a) that {\ttfamily FedMeS} is memory efficient, such that with storing as small as 30 data samples in local memory, an accuracy of $48\%$ can be achieved. As the memory size increases, the accuracy gradually improves, and persists around $55\%$ after the memory size reaches 150 data samples. For heterogeneous memory sizes across different clients is shown in \cref{mem here}.  As an instance, `Here10-50' denotes that every two clients are allocated memory samples spanning from 10 to 50, with a step of 10 samples, `Fix50' denotes a consistent allocation of 50 memory samples per client. Since each client is assigned with a unique task sequence, the memory sizes for each task is also heterogeneous. The $Acc\_Task$ boxplot reveals that {\ttfamily FedMeS} exhibits a broader accuracy range for each task under heterogeneous memory settings as opposed to fixed ones. Corroborating with  \cref{Time and ACC}, shown that performance in heterogeneous memory conditions is primarily influenced by the minimal memory sample size in each configuration. In terms of $Acc\_ALL$, it consistently surpasses other baselines from results in \cref{acc and FR 1}, further showing the robustness of {\ttfamily FedMeS} under various conditions.

{\bf Communication and computation efficiency.} We compare the communication and computation efficiency between {\ttfamily FedMeS} and {\ttfamily FedWeIT} in wall clock time, imposing a communication bandwidth limit of 40Mbps between the server and clients. The results, depicted in \cref{Time and ACC} (b), show that for all considered experiment settings, {\ttfamily FedMeS} outperforms {\ttfamily FedWeIT} by achieving higher accuracy using shorter execution time. This advantage becomes particularly prominent when experimenting with larger numbers of clients and tasks.
\section{Conclusion}
This paper has presented the challenges associated with catastrophic forgetting and client drift in PFCL and proposed the {\ttfamily FedMeS} framework as a solution to these issues. {\ttfamily FedMeS} levearges a small reference memory in the local training process to replay knowledge from previous tasks to alleviate forgetting; and the same memory is also used for the inference process by applying KNN-based Gaussian inference to further improve model personalization capability. We thoroughly analyzed the convergence behavior of {\ttfamily FedMeS}, and performed extensive experiments over various PFCL tasks. For all experiments, {\ttfamily FedMeS} uniformly outperforms existing techniques in terms of prediction accuracy and forgetting rate. 






\newpage
\appendix
\section*{Appendix}

\section{Detailed Theoretical Analysis}\label{sec:convergence}
In this section, we give theoretical analysis on how our algorithm uses local memories and provide convergence proof of the local training process.

\subsection{Reasoning of Local Continual Learning}\label{Reasoning of Local Continual Learning}

In this section, we provide inference on how \eqref{PFCL formal} can utilize memories of past tasks and how the update rule in \eqref{w_k update rule 1} and \eqref{w_k update rule 2} can solve the optimization problem.

To conduct continual learning in the centralized setting,  \cite{lopez2017gradient} raises the concept of \textit{episodic memory}, where for every task $t$, a central memory $\mathcal{M}_t$ is utilized to store examples sampled from task $t$. Denote $l(h_w; \mathcal{D})$ as the loss function hypothesis $h_w$ on dataset $\mathcal{D}$, where $h_w$ is the hypothesis with parameter $w$, and to avoid catastrophic forgetting, the authors raises that the loss of current model on previous tasks $j=1,...,t-1$, i.e. $l(h_w; \mathcal{M}_j)$ should not increase but decrease while minimizing loss on current task $t$, i.e. $l(h_w; \mathcal{D}^t)$, where $\mathcal{D}^t$ is task $t$'s dataset. However, this method requires computing on all memory block $\mathcal{M}_j$ for every $j<t$, which is computationally expensive.

Authors of \cite{chaudhry2018efficient}, simplify the above decreasing constraint of losses on previous tasks into into an inequality constraint: by every training step the \textit{average} memory loss over the previous tasks does not increase. Formally speaking, while learning task $t$, the objective is, 

\begin{equation}
\begin{aligned}
\text{minimize}_w\quad &l(h_w^t; \mathcal{D}^t) \\
\quad \text{s.t. } \quad&l(h^t_w; \mathcal{M}^t) \leq l(h_w^{t-1}; \mathcal{M}^t) \\
\quad \text{where } \quad&\mathcal{M}^t = \cup_{j<t}\mathcal{M}_j
\label{a-gem obj} 
\end{aligned}
\end{equation}

where $\mathcal{D}^t$ is dataset of current task $t$ and $h_w^t$ is the predictor state at the end of learning of task $t$. In the federated setting, we focus on the single client $k$ for task $t$, and the objective changes into,

\begin{equation}
    \min _{w_k^t} \quad \mathcal{L}\left(w_k^{t}; \mathcal{D}_k^t\right) \quad \text{s.t.} \quad  \mathcal{L}\left(w_k^{t}; \mathcal{M}_{k}^t\right) \leq \mathcal{L}\left(w_k^{t-1}; \mathcal{M}_{k}^t\right)
\end{equation}

where $w_k^t$, $\mathcal{D}_k^t$, $\mathcal{M}_k^t$ are respectively the model parameters, the dataset and the memory block of task $t$ on client $k$, as defined in Chapter \ref{Problem Definition} and \ref{Mmeory Design}. 

Moreover, \cite{lopez2017gradient} raises that it is unnecessary to store all the old predictors $h_w^j, j<t$ under the circumstance that the loss on the previous memories does not increase after each parameter update $g$. As the memorized examples being representative of past tasks, the increases of previous memory could be diagnosed by computing the angle between the previous loss gradient vector and the current proposed update. Therefore, \eqref{a-gem obj} could be restated as:

\begin{equation}
\left \langle g, g_{\operatorname{mem}} \right \rangle := \left \langle \frac{\partial l(h_w^t; \mathcal{D}^t)}{\partial w},\frac{\partial l(h_w^t; \mathcal{M}^t)}{\partial w} \right \rangle \geq 0
\label{a-gem gra} 
\end{equation}

where $g_{\operatorname{mem}}$ is the gradient computed using a batch randomly sampled from the memory, i.e. $(\mathbf{x}_{\operatorname{mem}}, y_{\operatorname{mem}}) \sim \mathcal{M}$. If the constraint in \eqref{a-gem gra} is satisfied, the current proposed update $g$ would decrease the loss on previous memories. However, if violation occurs, we could propose to project the proposed gradient $g$ to the closet gradient $\tilde{g}$ (in squared $l_2$ norm) satisfying \eqref{a-gem gra}. It can be rephrased as:

\begin{equation}
\text{minimize}_{\tilde{g}} \quad \frac{1}{2}\Vert g-\tilde{g}\Vert_2^2 \quad \text{s.t.} \quad \tilde{g}^\top g_{\operatorname{mem}} \geq 0
\label{a-gem l2} 
\end{equation}

In \cite{chaudhry2018efficient}, the optimization problem in \eqref{a-gem l2} is solved as below. With $z$ replacing $\tilde{g}$ and discard the term $g^\top g$, \eqref{a-gem l2} could be rewritten as below:

\begin{equation}
\text{minimize}_{z} \quad \frac{1}{2}z^\top z-g^\top z \quad \text{s.t.} \quad -z^\top g_{\operatorname{mem}} \leq 0
\label{a-gem z-replacing} 
\end{equation}

Note that the sign of inequality constraint is changed for Lagrangian dual problem. By adding the Lagrangian multiplier $\lambda$, the Lagrangian of \eqref{a-gem z-replacing} is:

\begin{equation}
\mathcal{L}(z, \lambda) = \frac{1}{2}z^\top z-g^\top z-\lambda z^\top g_{\operatorname{mem}}
\label{a-gem L} 
\end{equation}

The dual problem of \eqref{a-gem L} is:

\begin{equation}
\Theta(\lambda) = min_z \mathcal{L}(z, \lambda)
\label{a-gem dual} 
\end{equation}

The optimal value $z^*$ for the primal problem could be yielded by setting $\nabla_zL(z, \lambda)=0$, and we have: 
\begin{equation}
z^* = g+\lambda g_{\operatorname{mem}}
\label{a-gem z*} 
\end{equation}
Putting the value of $z^*$ back to dual problem \eqref{a-gem dual}, and the dual could be rewritten as:

\begin{equation}
\begin{aligned}
\Theta(\lambda) & = \frac{1}{2}(g+\lambda g_{\operatorname{mem}})\top(g+\lambda g_{\operatorname{mem}})-\\
& \quad\quad\quad\ g^\top(g+\lambda g_{\operatorname{mem}})-\lambda(g+\lambda g_{\operatorname{mem}})^\top g_{\operatorname{mem}}\\
& = \frac{1}{2}g\top g+\lambda g^\top_{\operatorname{mem}}g+\frac{1}{2}\lambda^2g^\top_{\operatorname{mem}}g_{\operatorname{mem}}-\\
& \quad\quad\quad g^Tg-\lambda g^\top g_{\operatorname{mem}}-\lambda g^\top g_{\operatorname{mem}}-\lambda^2g^\top_{\operatorname{mem}}g_{\operatorname{mem}}\\
& = -\frac{1}{2}g^\top g-\frac{1}{2}\lambda^2g^\top_{\operatorname{mem}}g_{\operatorname{mem}}-\lambda g^\top g_{\operatorname{mem}}
\label{a-gem nabla-lambda}
\end{aligned}
\end{equation}

The solution $\lambda^*$ is given by $\nabla_\lambda \Theta(\lambda) = 0$, i.e. $g^\top _{\operatorname{mem}}g_{\operatorname{mem}} \lambda +g^\top g_{\operatorname{mem}} = 0$. As a result,

\begin{equation}
\begin{aligned}
\lambda^* = -\frac{g^\top g_{\operatorname{mem}}}{g^\top _{\operatorname{mem}}g_{\operatorname{mem}}}
\label{a-gem lambda*}
\end{aligned}
\end{equation}

Putting back $\lambda^*$ in \eqref{a-gem z*}, the update rule would be:
\begin{equation}
\begin{aligned}
z^* = g-\frac{g^\top g_{\operatorname{mem}}}{g^\top _{\operatorname{mem}}g_{\operatorname{mem}}}g_{\operatorname{mem}} = \tilde{g}
\label{a-gem tildeg}
\end{aligned}
\end{equation}

 Next, we go back to the definition of gradient $g$ and $g_{\operatorname{mem}}$. $g$ is the proposed gradient update on the current task, and in the federated setting, it refers to the gradient of the personalized parameter $w_k^t$ on the client $k$'s dataset of task $t$, i.e.  $\mathcal{D}_k^t$. $g_{\operatorname{mem}}$ is the gradient from the previous tasks computed from a random subset of the memory, and in the federated setting, it refers to the gradient of the personalized parameter $w_k^t$ on the client $k$'s local memory till task $t$, i.e. $\mathcal{M}_k^t$. Formally we can define $g$ and $g_mem$ in PFCL problem as,
\begin{equation}
\begin{aligned}
g := & \nabla \mathcal{L}(w_k^t; \mathcal{D}_k^t) \\
g_{\operatorname{mem}} := & \nabla \mathcal{L}(w_k^t; \mathcal{M}_k^t)
\label{a-gem g and g_ref}
\end{aligned}
\end{equation}

Based on the above analysis, in a federated setting, the constraint \eqref{a-gem gra} turns into \eqref{nabla ineq} when we want to minimize $w_k^t$ in \eqref{PFCL formal} on client $k$ for current task $t$, 

$\bullet$ If $\left\langle\nabla \mathcal{L}(w_k^t;\mathcal{D}_k^{t}), \nabla \mathcal{L}(w_k^t;\mathcal{M}_k^t)\right\rangle \geq 0$, meaning that the angle of proposed gradient $g$ and gradient on the memory $g_{\operatorname{mem}} < 90^{\circ}$, so the update would not make for forgetting on the past examples. We could just calculate $\nabla \mathcal{L}(w_k; D_k^t)$ for this update, i.e.:
    \begin{equation}
        w_k^t \leftarrow  w_k^t-\eta_1 \nabla \mathcal{L}(w_k^t;\mathcal{D}_k^{t})
    \end{equation}
    Under this circumstance, the local update ensures the optimization both on the current task and local memories. As a result, the communication to the central server would not impede the local memories, and the regularization term $\lambda$ could be added to interpolate the global model and the local model, which turns into \eqref{w_k update rule 1}.
    
$\bullet$ Else, if $\left\langle\nabla \mathcal{L}(w_{k}^t;\mathcal{D}_k^{t}), \nabla \mathcal{L}(w_{k}^t;\mathcal{M}_k^t)\right\rangle < 0$, meaning that the angle of proposed gradient $g$ and gradient on the memory $g_{\operatorname{mem}} \geq 90^{\circ}$, so we should use the projected gradient $\tilde{g}$ for update, under which circumstance we have:
    \begin{equation}
    \begin{aligned}
    \tilde{g} & = g-\frac{g^\top g_{\operatorname{mem}}}{g^\top _{\operatorname{mem}}g_{\operatorname{mem}}}g_{\operatorname{mem}} \\
    & = \nabla \mathcal{L}(w_k^t; \mathcal{D}_k^t) - \\
& \quad\quad\quad \frac{\nabla \mathcal{L}(w_{k}^t; \mathcal{D}_k^{t})^{\top} \nabla \mathcal{L}(w_{k}^t;\mathcal{M}_k^t)}{\nabla \mathcal{L}(w_{k}^t;\mathcal{M}_k^t)^{\top} \nabla \mathcal{L}(w_{k}^t;\mathcal{M}_k^t)}\nabla \mathcal{L}(w_k^t; \mathcal{M}_k^t)
    \label{a-gem transform}
    \end{aligned}
    \end{equation}
    The corresponding update on current $w_k^t$ is:
    \begin{equation}
\begin{aligned}
w_k^t \gets  w_k^t-&\eta_2\bigg( \nabla \mathcal{L}(w_k^t; \mathcal{D}_k^t)- 
\\ &\frac{\nabla \mathcal{L}(w_k^t; \mathcal{D}_k^{t})^{\top} \nabla \mathcal{L}(w_k^t;\mathcal{M}_k^t)}{\nabla \mathcal{L}(w_k^t;\mathcal{M}_k^t)^{\top} \nabla \mathcal{L}(w_k^t;\mathcal{M}_k^t)}\nabla \mathcal{L}(w_k^t; \mathcal{M}_k^t) \bigg )
     \label{a-gem alpha2 local update}
\end{aligned}
\end{equation}
    which is exactly how the iteration takes place in \eqref{w_k update rule 2}.

Above is how the clients use local memory data to strengthen the personalized continual learning in PFCL.

\subsection{Proof of Theorem \cref{theroem converge to *}}
\label{Convergence Analysis of the Local Training Process}

\textit{Proof.} The proof follows the convergence analysis for Stochastic Gradient Descent in \cite{bottou2018optimization}.

We denote $g(w_k^{(s)}; \zeta_k^s)$ as $\nabla \mathcal{L}(w_k^{(s)}; \zeta_k^s)$. As the variance of $g(w_k^{(s)}; \zeta_k^s)$ can be written as,

\begin{equation}
\begin{aligned}
     \mathbb{V}_{\zeta_k^s}[g(w_k^{(s)}; \zeta_k^s)] := \mathbb{E}_{\zeta_k^s}&[\Vert g(w_k^{(s)}; \zeta_k^s)\Vert^2_2]-\\
& \Vert\mathbb{E}_{\zeta_k^s}[g(w_k^{(s)}; \zeta_k^s)]\Vert_2^2
\end{aligned}
\end{equation}

According to Assumptions \ref{assumption 4} and \ref{assumption 5},
\begin{equation}
\begin{aligned}
    \mathbb{E}_{\zeta_k^s}[\Vert g(w_k^{(s)}; \zeta_k^s)\Vert^2_2] &= \mathbb{V}_{\zeta_k^s}[g(w_k^{(s)}; \zeta_k^s)]-\\
&\quad\quad\quad\quad\quad\quad\Vert\mathbb{E}_{\zeta_k^s}[g(w_k^{(s)}; \zeta_k^s)]\Vert_2^2\\
    &\leq M+M_V\Vert\nabla\mathcal{L}(w_k^{(s)}; \mathcal{M}_k^t)\Vert^2_2+ \\
& \quad\quad\quad\quad\quad\quad \mu_G^2\Vert\nabla\mathcal{L}(w_k^{(s)}; \mathcal{M}_k^t)\Vert_2^2\\
    & \leq M+M_G\Vert\mathcal{L}(w_k^{(s)}; \mathcal{M}_k^t)\Vert_2^2 \quad \\
&\text{with} \ M_G := M_V+\mu_G^2 \geq \mu^2 > 0
\end{aligned}
\end{equation}

The update rule of \eqref{w_k update rule 2} comes with the constraint problem of \eqref{a-gem l2}, and \eqref{a-gem transform} \eqref{a-gem alpha2 local update} demonstrates that,
\begin{equation}
    w_k^{(s+1)} = w_k^{(s)} -\eta_2\tilde{g}\quad \text{where} \quad \tilde{g}^\top g(w_k^{(s)}; \zeta_k^s) \geq 0
\end{equation}

Since $\tilde{g}^\top g(w_k^{(s)}; \zeta_k^s) \geq 0$, suppose $g(w_k^{(s)}; \zeta_k^s) \neq 0$ and reversible, then there exists a semi-positive definite matrix $A_s$,
\begin{equation}
    \tilde{g} = A_sg(w_k^{(s)}; \zeta_k^s)
\end{equation}

According to Assumption \ref{assumption 1-3} that $\mathcal{L}(w)$ is $L$-smooth,
\begin{equation}
\begin{aligned}
 &   \mathcal{L}(w_k^{(s+1)}; \mathcal{M}_k^t) -  \mathcal{L}(w_k^{(s)}; \mathcal{M}_k^t) \\
& \leq \nabla\mathcal{L}(w_k^{s}; \mathcal{M}_k^t)^\top\left(w_k^{(s+1)}-w_k^{(s)}\right)+\\
&\quad\quad\quad\quad\quad\quad\quad\quad\quad\quad\quad \frac{1}{2}L\Vert w_k^{(s+1)}-w_k^{(s)}\Vert^2\\
    &  = -\nabla\mathcal{L}(w_k^{(s)}; \mathcal{M}_k^t)^\top\eta_2A_sg(w_k^{(s)}; \zeta_k^s)+ \\
& \quad\quad\quad\quad\quad\quad\quad\quad\quad\quad\quad\frac{1}{2}L\eta_2^2A_s^\top A_s\Vert g(w_k^{(s)}; \zeta_k^s)\Vert^2_2\\
\end{aligned}
\end{equation}

Both sides of the equation simultaneously find the expectation for $g(w_k^s; \zeta_k^s)$, 
\begin{equation}
\begin{aligned}
    &\mathbb{E}_{\zeta_k^s}[\mathcal{L}(w_k^{(s+1)}; \mathcal{M}_k^t)] -  \mathcal{L}(w_k^{(s)}; \mathcal{M}_k^t)\\
    &\leq -\eta_2A\nabla\mathcal{L}(w_k^{(s)}; \mathcal{M}_k^t)^\top\mathbb{E}_{\zeta_k^s}[g(w_k^{(s)}; \zeta_k^s)]+\\
& \quad\quad \frac{1}{2}\eta_2^2LA_s^\top A_s\mathbb{E}_{\zeta_k^s}[\Vert g(w_k^{(s)}; \zeta_k^s)\Vert^2_2]\\
    &\leq-\eta_2\mu A_s\Vert\nabla\mathcal{L}(w_k^{(s)}; \mathcal{M}_k^t)\Vert_2^2+\\
& \quad\quad \frac{1}{2}\eta^2_2LA_s^\top A_s(M+M_G\Vert\nabla\mathcal{L}(w_k^{(s)}; \mathcal{M}_k^t)\Vert_2^2)\\
    & = (-\mu-\frac{1}{2}\eta_2LM_GA_s^\top)\eta_2A_s\Vert\nabla\mathcal{L}(w_k^{(s)}; \mathcal{M}_k^t)\Vert_2^2+\\
& \quad\quad \frac{1}{2}\eta_2^2LMA_s^\top A_s
\end{aligned}
\end{equation}

Assume that the learning rate $\eta_2$ could be adjusted in every iteration $s$, denoted as $\eta_s$ for simplicity, and follows the below constraint,
\begin{equation}
    \eta_s \leq \frac{\mu}{LM_GA_s}
    \label{s assumption}
\end{equation}

With Assumption \ref{assumption 1-3} that $\mathcal{L}(w)$ is $c$-strongly convex, and with \eqref{s assumption},
\begin{equation}
\begin{aligned}
        &\mathbb{E}_{\zeta_k^s}[\mathcal{L}(w_k^{(s+1)}; \mathcal{M}_k^t)] -  \mathcal{L}(w_k^{(s)}; \mathcal{M}_k^t) \\
&\leq  \frac{1}{2}\eta_s\mu A_s \Vert\nabla\mathcal{L}(w_k^{(s)}; \mathcal{M}_k^t)\Vert_2^2+\frac{1}{2}\eta_s^2LMA_s^\top A_s\\
        &\leq -\eta_sc\mu A_s\left(\mathcal{L}(w_k^{(s)}; \mathcal{M}_k^t)-\mathcal{L}^*(
        \mathcal{M}_k^t)\right)+\\
&\quad\quad\quad\quad\quad\quad\quad\quad\quad\quad\quad\quad\quad\quad \frac{1}{2}\eta_s^2LMA_s^\top A_s
    \end{aligned}
\end{equation}

where $\mathcal{L}^*(\mathcal{M}_k^t) = \min \mathcal{L}(w_k; \mathcal{M}_k^t)$. Add $\mathcal{L}(w_k^{(s)}; \mathcal{M}_k^t)-\mathcal{L}^*(
        \mathcal{M}_k^t)$ on both sides,
\begin{equation}
\begin{aligned}
    &\mathbb{E}_{\zeta_k^s}[\mathcal{L}(w_k^{(s+1)}; \mathcal{M}_k^t)] - \mathcal{L}^*(\mathcal{M}_k^t) \leq\\ 
&\left(1-\eta_sc\mu A_s\right)\left(\mathcal{L}(w_k^{(s)}; \mathcal{M}_k^t)-\mathcal{L}^*(\mathcal{M}_k^t)\right)+\\
&\quad \quad \quad \quad \quad \quad \quad \quad \quad \quad \quad \quad \quad \quad \frac{1}{2}\eta_s^2LMA_s^\top A_s\\
    &\mathbb{E}_{\zeta_k^s}[\mathcal{L}(w_k^{(s+1)}; \mathcal{M}_k^t) - \mathcal{L}^*(\mathcal{M}_k^t)] \leq\\ 
&\left(1-\eta_sc\mu A_s\right)\mathbb{E}[\mathcal{L}(w_k^{(s)}; \mathcal{M}_k^t)-\mathcal{L}^*(\mathcal{M}_k^t)]+\\
&\quad \quad \quad \quad \quad \quad \quad \quad \quad \quad \quad \quad \quad \quad \frac{1}{2}\eta_s^2LMA_s^\top A_s\\
    \label{L inequaility}
\end{aligned}
\end{equation}

To prove Theorem \ref{theroem converge to *}, we prove below inequality for $s\in\mathbb{N}^+$ by induction first,
\begin{equation}
\begin{aligned}
     \mathbb{E}_{\zeta_k^s}&[\mathcal{L}(w_k^{(s+1)}; \mathcal{M}_k^t) - \mathcal{L}^*(\mathcal{M}_k^t)] \leq \\
&\left(\prod_{i=1}^s(1-\eta_ic\mu A_i)\right)\mathbb{E}[\mathcal{L}(w_k^{(1)}; \mathcal{M}_k^t)-\mathcal{L}^*(\mathcal{M}_k^t)]+\\
&\quad\quad\quad\quad\quad\quad\quad\quad\quad\quad\quad\quad\quad\quad\quad\quad\quad\quad \frac{LM}{2c^2\mu^2}
     \label{induction ref}
\end{aligned}
\end{equation}
When $s=1$,
\begin{equation}
\begin{aligned}
    \mathbb{E}_{\zeta_k^1}&[\mathcal{L}(w_k^{(2)}; \mathcal{M}_k^t) - \mathcal{L}^*(\mathcal{M}_k^t)]  \leq\\
& \left(1-\eta_1c\mu A_1\right)\mathbb{E}[\mathcal{L}(w_k^{(1)}; \mathcal{M}_k^t)-\mathcal{L}^*(\mathcal{M}_k^t)]+\\
&\quad\quad\quad\quad\quad\quad\quad\quad\quad\quad\quad\quad\quad \frac{LM\eta_1^2A_1^\top A_1}{2}
    \label{s=1}
\end{aligned}
\end{equation}

With \eqref{s assumption}, further for every $s\in \mathbb{N}^+$,
\begin{equation}
    \eta_sc\mu A_s \leq \frac{\mu}{LM_GA_s}\cdot c\mu A_s = \frac{c\mu^2}{LM_G^2} \leq\frac{c\mu^2}{L\mu^2}=\frac{c}{L} \leq 1
    \label{eta<1}
\end{equation}

which in turn,
\begin{equation}
    \eta_sA_s \leq \frac{1}{c\mu}
    \label{eta_sA_s}
\end{equation}

Take \eqref{eta_sA_s} into \eqref{s=1},
\begin{equation}
\begin{aligned} 
    &\mathbb{E}_{\zeta_k^1}[\mathcal{L}(w_k^{(2)}; \mathcal{M}_k^t) - \mathcal{L}^*(\mathcal{M}_k^t)]  \\
&\leq\left(1-\eta_1c\mu A_1\right)\mathbb{E}[\mathcal{L}(w_k^{(1)}; \mathcal{M}_k^t)-\mathcal{L}^*(\mathcal{M}_k^t)]+\\
&\quad\quad\quad\quad\quad\quad\quad\quad\quad\quad\quad\quad\quad\quad\quad\quad \frac{LM\eta_1^2A_1^\top A_1}{2}\\ 
    & \leq \left(1-\eta_1c\mu A_1\right)\mathbb{E}[\mathcal{L}(w_k^{(1)}; \mathcal{M}_k^t)-\mathcal{L}^*(\mathcal{M}_k^t)]+\frac{LM}{2c^2\mu^2}
\end{aligned}
\end{equation}
which meets \eqref{induction ref} when $s=1$. Assume that \eqref{induction ref} holds for $s$, then, in $s+1$ iteration,

\begin{equation}
    \begin{aligned}
        &\mathbb{E}_{\zeta_k^{s+1}}[\mathcal{L}(w_k^{(s+2)}; \mathcal{M}_k^t) - \mathcal{L}^*(\mathcal{M}_k^t)] \\
        & \leq \left(1-\eta_{s+1}c\mu A_{s+1}\right)\mathbb{E}[\mathcal{L}(w_k^{(s+1)}; \mathcal{M}_k^t)-\mathcal{L}^*(\mathcal{M}_k^t)]+\\
&\quad\quad\quad\quad\quad\quad\quad\quad\quad\quad\quad\quad u\frac{1}{2}\eta_{s+1}^2LMA_{s+1}^\top A_{s+1}\\
        & \leq \left(1-\eta_{s+1}c\mu A_{s+1}\right)*\\
&\quad\left(\mathcal{N} \mathbb{E}[\mathcal{L}(w_k^{(1)}; \mathcal{M}_k^t)-\mathcal{L}^*(\mathcal{M}_k^t)]+\frac{LM}{2c^2\mu^2}\right)+\\
&\quad\quad\quad\quad\quad\quad\quad\quad\quad\quad\quad\quad\frac{1}{2}\eta_{s+1}^2LMA_{s+1}^\top A_{s+1}\\
        & = \mathcal{N}\mathbb{E}[\mathcal{L}(w_k^{(1)}; \mathcal{M}_k^t)-\mathcal{L}^*(\mathcal{M}_k^t)]+\\
&\quad\left(1-\eta_{s+1}c\mu A_{s+1}\right)\frac{LM}{2c^2\mu^2}+\frac{1}{2}\eta_{s+1}^2LMA_{s+1}^\top A_{s+1}\\
        & \leq \mathcal{N}\mathbb{E}[\mathcal{L}(w_k^{(1)}; \mathcal{M}_k^t)-\mathcal{L}^*(\mathcal{M}_k^t)]+\\
&\quad\left(1-\eta_{s+1}c\mu A_{s+1}\right)\frac{LM}{2c^2\mu^2}+\frac{\eta_{s+1}LMA_{s+1}}{2cu}\\
        & = \mathcal{N}\mathbb{E}[\mathcal{L}(w_k^{(1)}; \mathcal{M}_k^t)-\mathcal{L}^*(\mathcal{M}_k^t)]+\\
&\quad\frac{LM}{2c^2\mu^2}-\frac{\eta_{s+1}LMA_{s+1}}{2c\mu}+\frac{\eta_{s+1}LMA_{s+1}}{2cu}\\
        & = \mathcal{N}\mathbb{E}[\mathcal{L}(w_k^{(1)}; \mathcal{M}_k^t)-\mathcal{L}^*(\mathcal{M}_k^t)]+\frac{LM}{2c^2\mu^2}
    \end{aligned}
\end{equation}
where $\mathcal{N}=\left(\prod_{i=1}^{s+1}(1-\eta_ic\mu A_i)\right)$. This demonstrates that \eqref{induction ref} also holds for $s+1$, completing the proof of \eqref{induction ref}.

According to \eqref{eta<1},
\begin{equation}
\begin{aligned}
    &\mathbb{E}[\mathcal{L}(w_k^{(s)}; \mathcal{M}_k^t)-\mathcal{L}^*(\mathcal{M}_k^t)]\\
 &\leq \left(\prod_{i=1}^s(1-\eta_ic\mu A_i)\right)\mathbb{E}[\mathcal{L}(w_k^{(1)}; \mathcal{M}_k^t)-\mathcal{L}^*(\mathcal{M}_k^t)]+\\
&\quad\quad\quad\quad\quad\quad\quad\quad\quad\quad\quad\quad\quad\quad\quad\quad\quad\quad\quad \frac{LM}{2c^2\mu^2}\\
    &\stackrel{s\rightarrow \infty}{\longrightarrow}   \frac{LM}{2c^2\mu^2}
\end{aligned}
\end{equation}
completing the proof for Theorem \ref{theroem converge to *}.

\textbf{Discussions.} Theorem \ref{theroem converge to *} shows that $\mathcal{L}(w_k^t; \mathcal{M}_k^t)$ can always converge to $\mathcal{L}^*(\mathcal{M}_k^t)$ with the update rule of \eqref{w_k update rule 2}. Thus, whenever the current parameter $w_k^t$ violates the inequality limit \eqref{nabla ineq} when running the update rule of \eqref{w_k update rule 1}, by applying \eqref{w_k update rule 2} repeatedly the loss on memory $\mathcal{L}(w_k^t; \mathcal{M}_k)$ would always converge to the optimal $\mathcal{L}^*(\mathcal{M}_k)$. As $\mathcal{L}(w_k^{(t-1)}; \mathcal{M}_k)$ can be viewed as a constant when solving task $t$, under the assurance of the convergence analysis, updating rule \eqref{w_k update rule 2} can ensure that current loss on $\mathcal{M}_k$ would gradually get smaller and finally, at a certain step, satisfy the constraint of \eqref{change d_k to m_k}, and the algorithm goes on. 

\subsection{Proof of Theorem \ref{theorem ditto local convergence-mainbody}}
\label{Proof of Theorem 2}

The proof of Theorem \ref{theorem ditto local convergence-mainbody} can be referred to Theorem 10 in \cite{li2021ditto}.

\label{last-page}
\end{multicols}
\label{last-page}
\end{document}